%% file: root.tex
\documentclass[letterpaper, 10 pt, conference]{ieeeconf}  
\IEEEoverridecommandlockouts                              
\usepackage{caption}        
\usepackage{xcolor}
\let\labelindent\relax
\usepackage{enumitem}
\long\def\invis#1{}
\usepackage [english]{babel}
\usepackage [autostyle, english = american]{csquotes}
\MakeOuterQuote{"}
\usepackage{booktabs}
\usepackage[colorlinks]{hyperref}
\usepackage{wrapfig}
\usepackage{amssymb}

\usepackage{subcaption}
\usepackage{amsmath}
\usepackage{amssymb}
\usepackage{algorithm}
\usepackage{algorithmic}
\usepackage{float}

\usepackage{bbm}

\usepackage{graphicx}
\usepackage{cite}
\usepackage[font=small,skip=2pt]{caption} 
\usepackage{amsmath}
\usepackage{siunitx}
\usepackage{float}

\makeatletter
\renewcommand\paragraph{\@startsection{paragraph}{4}{\z@}%
            {-2.5ex\@plus -1ex \@minus -.25ex}%
            {1.25ex \@plus .25ex}%
            {\normalfont\normalsize\bfseries}}
\makeatother

\usepackage{xcolor}

\title{\LARGE \bf
\textsc VARP: Reinforcement Learning from Vision-Language Model Feedback with Agent Regularized Preferences }

\author{Anukriti Singh$^{*1}$, Amisha Bhaskar$^{*1}$, Peihong Yu$^{1}$, Souradip Chakraborty$^{1}$, Ruthwik Dasyam$^{1}$,\\
Amrit Bedi$^{2}$, Pratap Tokekar$^{1}$% <-this % stops a space
\thanks{*Both the authors contributed equally}%
\thanks{$^{1}$University of Maryland, College Park.}%
\thanks{$^{2}$Department of Computer Science, University of Central Florida.}%
\thanks{Correspondence to: Amisha Bhaskar (\texttt{amishab@umd.edu}) and Anukriti Singh (\texttt{anukriti@umd.edu}) }%
\thanks{This work is supported by the National Science Foundation under Grant No. 1943368 and an Amazon Research Award.}}

\makeatletter\def\@bibliographyfile{references}\makeatother
\begin{document}

\maketitle
\thispagestyle{empty}
\pagestyle{empty}

%%%%%%%%%%%%%%%%%%%%%%%%%%%%%%%%%%%%%%%%%%%%%%%%%%%%%%%%%%%%%%%%%%%%%%%%%%%%%%%%

\input{sections/0.abstract}
\input{sections/1.introduction}

\input{sections/2.related_work}
\input{sections/3.problem_statement}

\input{sections/5.system_overview}

\input{sections/6.experiments}
\input{sections/7.conclusion}

\bibliographystyle{IEEEtran}
\bibliography{IEEEabrv,references}

\end{document}

%% file: sections/0.abstract.tex
\begin{abstract}

Designing reward functions for continuous-control robotics often leads to subtle misalignments or reward hacking, especially in complex tasks. Preference-based RL mitigates some of these pitfalls by learning rewards from comparative feedback rather than hand-crafted signals, yet scaling human annotations remains challenging. Recent work uses Vision-Language Models (VLMs) to automate preference labeling, but a single final-state image generally fails to capture the agent’s full motion. In this paper, we present a two-part solution that both improves feedback accuracy and better aligns reward learning with the agent’s policy. First, we overlay \emph{trajectory sketches} on final observations to reveal the path taken, allowing VLMs to provide more reliable preferences—improving preference accuracy by approximately 15–20\% in metaworld tasks. Second, we \emph{regularize} reward learning by incorporating the agent’s performance, ensuring that the reward model is optimized based on data generated by the current policy; this addition boosts episode returns by 20–30\% in locomotion tasks. Empirical studies on metaworld demonstrate that our method achieves, for instance, around 70-80\% success rate in all tasks, compared to below 50\% for standard approaches. These results underscore the efficacy of combining richer visual representations with agent-aware reward regularization. Supplementary material can be found on our project \href{https://raaslab.org/projects/VARP/}{website}.

\end{abstract}

%% file: sections/1.introduction.tex
\section{Introduction}
\label{sec:introduction}

% Reinforcement Learning (RL) has achieved remarkable success in robotics, allowing agents to learn complex behaviors for tasks such as manipulation \cite{kilinc2021reinforcement,bhaskar2024planrl}, locomotion \cite{liu2020robot, terp}, and object rearrangement \cite{wu2022targf,szot2021habitat}. Yet, reward specification continues to be a central challenge \cite{knox2022reward,leike2018scalable}. Manually crafted rewards often fail to capture the nuances of real-world objectives, and even minor design errors can lead to unintended behaviors (often referred to as 'reward hacking' \cite{amodei2016concrete,skalse2022defining}). Although collecting expert demonstrations \cite{chan2014application, ferraguti2015bilateral, bimbo2017teleoperation} or engineering detailed cost functions can partly address this challenge, these solutions tend to be labor-intensive and susceptible to domain-specific biases \cite{knox2022reward}.

Reinforcement Learning (RL) has achieved remarkable success in robotics, enabling agents to learn complex behaviors for tasks such as manipulation \cite{kilinc2021reinforcement,bhaskar2024planrl}, locomotion \cite{liu2020robot, terp}, and object rearrangement \cite{wu2022targf,szot2021habitat}. Yet, reward specification continues to be a central challenge \cite{knox2022reward,leike2018scalable}. Hand-crafted rewards frequently fail to capture the nuances of real-world objectives, and even minor design errors can lead to unintended behaviors (often referred to as "reward hacking" \cite{amodei2016concrete,skalse2022defining}). Although methods such as Multi-Objective RL (MORL) \cite{mossalam2016multi} offers alternative formulations by considering multiple objectives simultaneously, they still require accurate reward signals to balance competing goals. In contrast, Learning from Human Feedback (RLHF) sidesteps explicit reward engineering by relying on comparative judgments, making it a compelling alternative for aligning rewards with human preferences.

\textbf{Reinforcement Learning from Human Feedback (RLHF)} \cite{lee2021pebble,christiano2017deep} replaces numeric rewards with human-generated comparisons. In typical RLHF, annotators select which of two trajectories better meets the task goals, providing a straightforward and intuitive way to capture performance differences \cite{lee2021pebble}. Although in principle one could extend these comparisons to rank multiple trajectories simultaneously, pairwise comparisons have proven to be a practical choice, as they simplify the annotation process and reduce the cognitive load on evaluators. Nevertheless, RLHF still suffers from the high cost of human labeling, motivating the recent adoption of automated feedback via Vision-Language Models (VLMs) \cite{wang2024rl,10445007}.

% \textbf{Reinforcement Learning from Human Feedback (RLHF)} \cite{lee2021pebble,christiano2017deep} offers a more direct approach by replacing numeric rewards with human-generated preferences. Instead of assigning absolute scores, human annotators choose which of two trajectories better meets the task goals \cite{lee2021pebble}. This method, however, still relies on extensive human involvement, an obstacle to large-scale deployment in robotics \cite{lee2021b}. Recent work alleviates this annotation burden by leveraging Vision-Language Models (VLMs) \cite{10445007}. These models using VLMs generate preference judgments based on image and text inputs, thereby reducing the need for real-time human supervision \cite{wang2024rl}.

Despite these advances, current VLM-based methods such as RL-VLM-F \cite{wang2024rl} often rely on a \emph{single final-state image} to represent an entire trajectory. In many robotic tasks, crucial information about the motion—such as the path efficiency or smoothness of intermediate steps—is lost when only the end state is considered. For example, two trajectories may lead to the same final configuration, yet one might follow a more direct and efficient path while the other takes a longer, more convoluted route. Focusing solely on the end state leaves the model blind to these critical differences. Moreover, although the VLM may initially provide accurate comparisons, optimizing the reward model only on such static data—without accounting for the evolving policy that generates diverse trajectories—can lead to suboptimal training. It is therefore essential to incorporate the agent’s performance throughout the trajectory, ensuring that the learned reward remains aligned with the most efficient and effective behavior.

% To address these issues, we introduce a novel pipeline called \textbf{VARP (VLM + Agent Regularized Preferences)}. First, rather than using only the final-state image, we overlay a 2D trajectory sketch onto the final observation to capture the entire motion sequence. This enriched visual representation significantly improves the VLM’s ability to generate reliable preference judgments. Second, and equally important, we introduce an \emph{agent-aware reward regularization scheme}. As the policy evolves, the distribution of states changes; if the reward model is optimized solely on static preference data, it may become misaligned with the agent's current performance. By incorporating the agent’s own returns into the reward-learning objective, our approach dynamically penalizes reward functions that yield low returns, thereby ensuring that the learned reward remains aligned with effective, real-time behaviors.

In this paper, we introduce a novel pipeline called \textbf{VARP (VLM + Agent Regularized Preferences)} that enhances preference-based RL by combining \emph{trajectory-aware feedback} \cite{gu2023rt,zhi2023learning,yu2025sketchtoskill} with an \emph{agent-aware reward regularization scheme} \cite{chakraborty2024parl}. First, rather than relying on a single final-state image and a textual prompt, we overlay 2D trajectory sketches on the final observation to capture the robot’s full motion. This enriched visual representation provides the VLM with crucial temporal context, enabling it to generate more reliable and precise preference judgments. Second, we address the challenge that static preference data can become misaligned as the policy evolves. If the reward model is optimized solely on such data, it may fail to reflect the current performance of the agent. To mitigate this, we integrate agent performance into the reward-learning objective, dynamically penalizing reward functions that yield low returns. This agent-aware regularization ensures that the learned reward remains aligned with effective, real-time behaviors, ultimately leading to more stable and robust policy improvement.

% In this paper, we introduce a novel pipeline called \textbf{VARP (VLM + Agent Regularized Preferences)}, which enhances preference-based RL by incorporating both \emph{trajectory-aware feedback}\cite{gu2023rt,zhi2023learning,yu2025sketchtoskill} and an \emph{agent-aware reward regularization scheme}\cite{chakraborty2024parl}. First, instead of providing the VLM with a single final-state image along with a textual prompt about the goal, we generate 2D sketches overlaid on the final image frame observation to illustrate the robot’s motion throughout the episode. This gives the model a richer understanding of how actions unfold, improving the quality of the preference judgments from the VLM. Second, we integrate agent preference into the reward-learning process itself. 

% \todo{by penalizing reward functions that produce low returns for the policy, we mitigate the risk of misalignment as the agent explores new states.}

In summary, our contributions are threefold:
\begin{itemize}
    \item We introduce a \textbf{trajectory-aware} preference labeling technique that leverages 2D sketches to enrich the VLM’s perception of a robot’s motion, leading to more precise comparisons.
    \item We propose \textbf{VARP}, a framework that balances external, sketch-based preference signals with the agent’s performance during reward optimization, ensuring that the learned reward remains aligned with the current policy.
    \item Through experiments on MetaWorld~\cite{yu2021metaworld} and DMControl~\cite{tassa2018deepmind} benchmarks, we demonstrate that VARP provides more accurate preference models and robust policy improvements.
\end{itemize}

\begin{figure*}[t!]
    \centering
    \makebox[\textwidth]{
        \includegraphics[width=1.1\textwidth]{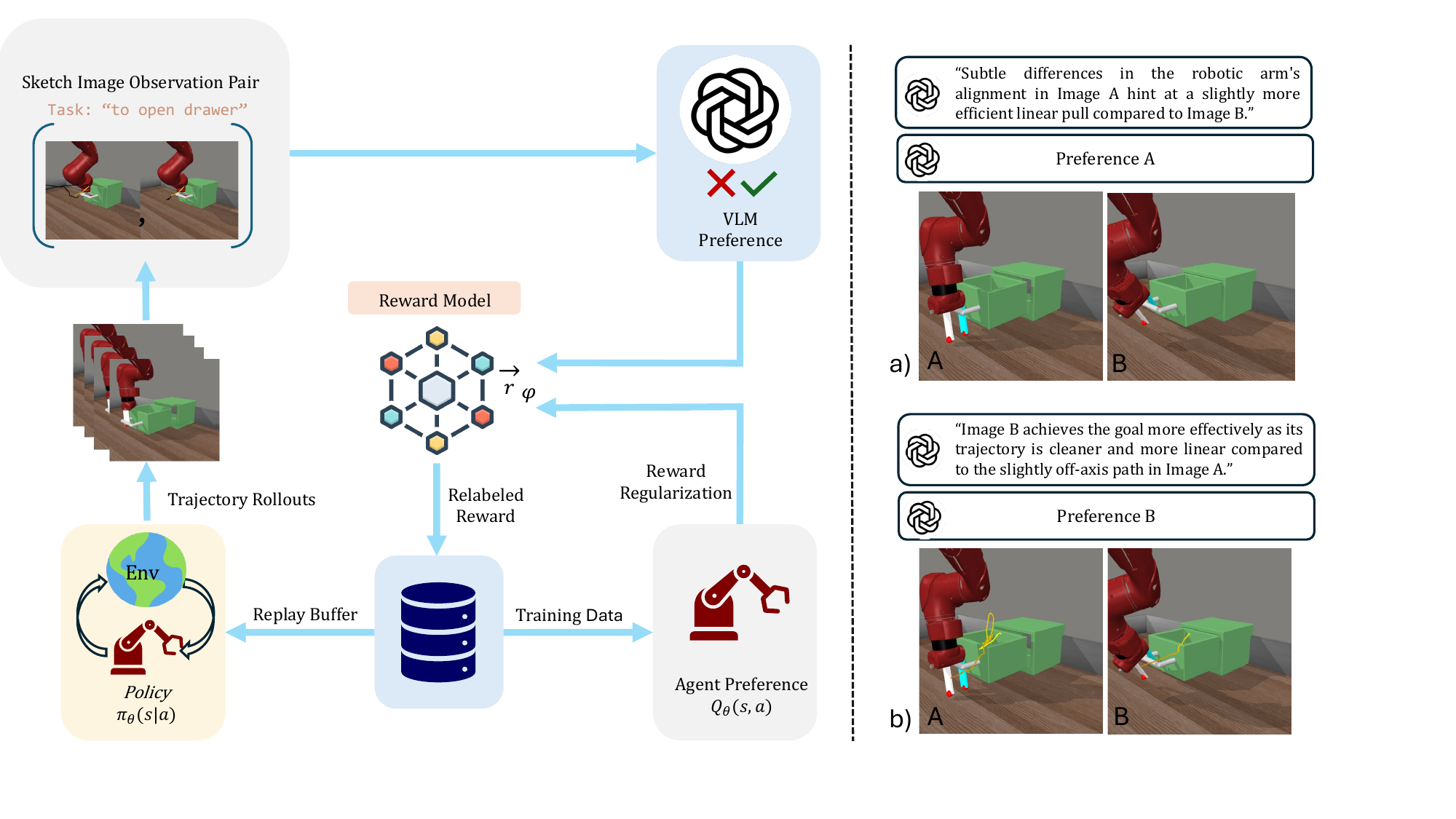}
    }
    \vspace{-3mm}
    \caption{
\textbf{Left:} This diagram breaks down our method into four key stages: (1) sketch data generation from the full trajectory, (2) two-stage VLM preference querying using the sketched observations, (3) reward model training that balances VLM feedback with agent performance, and (4) policy optimization using the learned reward.  This comprehensive approach enhances feedback accuracy and stabilizes policy learning. \textbf{Right:} Comparison of VLM preference outputs when using final-state images with (VARP) and without~\cite{wang2024rl} trajectory sketches. The added sketches provide crucial temporal context, resulting in more accurate preference judgments.}

    \label{fig:sketch_rl}
    \vspace{-5mm}
\end{figure*}

% By integrating richer visual cues and agent-centric reward regularization into a single pipeline, VARP offers a step toward scalable, human-aligned RL in settings that demand both nuanced feedback and consistent performance as policies evolve.

%% file: sections/2.related_work.tex
\section{RELATED WORK}
\label{section:related_work}

\textbf{Reward Learning and Human Feedback in Reinforcement Learning} 
Recent reinforcement learning research has tackled the challenge of aligning reward models with evolving policy behaviors and human preferences. Approaches including PEBBLE \cite{lee2021pebble} and RIME \cite{anonymous2024rime} address reward model generalization by using experience relabeling \cite{christiano2017deep} and de-noising discriminators \cite{anonymous2024rime}, respectively. However, these methods maintain a separation between reward learning and policy optimization, which can lead to misalignments. In robotics, preference-based learning has been further explored by Wilson, Fern, and Tadepalli \cite{NIPS2012_16c222aa} and through dueling bandits approaches as described by Dudík et al. \cite{dudik2015contextual} and Bengs et al. \cite{bengs2021preference}. Christiano et al. \cite{christiano2017deep} scaled preference-based methods to large-scale continuous control tasks despite sample inefficiencies associated with on-policy learning, issues partially mitigated by incorporating additional demonstrations \cite{ibarz2018reward} and non-binary preference rankings \cite{cao2020weak}. The Pebble framework \cite{lee2021pebble} advanced the field by integrating off-policy learning with pre-training while contending with reward model over-optimization. In contrast, VARP integrates reward learning and policy optimization via a first-order bilevel optimization approach that employs trajectory-level regularization supported by sketch-augmented vision-language feedback. This integrated strategy dynamically balances human and agent preferences while penalizing deviations from high-return states to prevent reward hacking and ensure alignment with task objectives.

% \textbf{Human Feedback in RL.}
% Preference-based learning in robotics offers a promising avenue for deriving rewards that reflect human or expert preferences. Wilson, Fern, and Tadepalli (2012) \cite{NIPS2012_16c222aa}. Dueling bandits, such as those discussed by Dudík et al. (2015) \cite{dudik2015contextual} and Bengs et al. (2021)\cite{bengs2021preference}, attempt to minimize regret through pairwise comparisons, yet their practical application in robotics remains challenging due to their theoretical nature. 
% Christiano et al. (2017)\cite{christiano2017deep}  scaled preference-based methods to large-scale continuous control tasks, though they faced sample inefficiency issues associated with on-policy learning, which were later partially mitigated by incorporating additional demonstrations \cite{ibarz2018reward} and non-binary preference rankings \cite{cao2020weak}. The Pebble framework \cite{lee2021pebble} further advanced this field by integrating off-policy learning with pre-training, although it still grapples with challenges such as reward model over-optimization, which can undermine performance in complex environments.

\textbf{Vision-Language Model Feedback.}
Recent advancements harness large VLMs to automatically generate preference labels, significantly reducing the reliance on costly human annotations \cite{wang2024rl}. However, using single images often introduces ambiguity, particularly in tasks where visual states may appear similar but represent unsafe or inefficient trajectorie. To tackle this, our approach, VARP, introduces 2D trajectory sketches that supplement final images, providing VLMs with crucial temporal context. This addition allows VLMs to effectively parse both spatial patterns and motion dynamics, enhancing their ability to discern subtle differences in agent behavior akin to human evaluative processes. Similar approaches have explored leveraging pre-trained models for direct reward function generation in various domains \cite{kwon2023reward,yu2023language}, but often struggle with the specificity and noise in task alignment—challenges that VARP addresses by enhancing visual feedback with clear, actionable content.

% % \vspace{}
% \textbf{Regularizing Learned Rewards.}
% Traditional approaches like PEBBLE and RIME have addressed the issue of reward models not generalizing as policy behaviors evolve, using experience relabeling and de-noising discriminators respectively \cite{christiano2017deep, anonymous2024rime}. However, these methods often separate the processes of reward learning and policy optimization, which can lead to misalignments. VARP innovatively integrates these processes, applying a first-order, bilevel optimization approach that uses trajectory-level regularization supported by sketch-augmented VLM feedback. This not only balances human and agent preferences dynamically but also penalizes deviations from high-return states. Such an approach prevents "reward hacking," where agents exploit poorly specified rewards, ensuring the learned policies truly align with task objectives.

%% file: sections/3.problem_statement.tex
\section{Preliminaries}
\label{sec:problem}

We consider a task modeled as a Markov Decision Process (MDP) defined by $ M = (S, A, P, r, \gamma)$,
where $S$ is the state space, $A$ is the action space, $P(s' \mid s, a)$ is the transition function, $r: S \times A \to \mathbb{R}$ is the (unknown) reward function, and $\gamma \in (0,1)$ is the discount factor. Let $\pi_\theta(a \mid s)$ be the agent's policy, parameterized by $\theta$, seeking to maximize the expected return
\[
  R(\pi_\theta) \;=\; \mathbb{E}_{\substack{s_t \sim P \\ a_t \sim \pi_\theta}} \Bigl[\sum_{t=0}^{H-1} \gamma^t \, r(s_t, a_t)\Bigr].
\]

\subsection{Preference-Based Reward Learning}

Instead of directly observing \(r(\cdot)\), we assume access to \emph{pairwise preferences} over trajectory segments. A \emph{trajectory segment} \(\tau = (s_0, a_0, s_1, a_1, \dots, s_{k})\) is a finite rollout under some policy. Given two segments, \(\tau^a\) and \(\tau^b\), a label \(y \in \{0,1\}\) indicates which segment is preferred. We adopt the standard Bradley-Terry model~\cite{Bradley1952RankAO}:
\begin{equation}
  P(\tau^a \succ \tau^b) \;=\;
  \frac{\exp\!\Bigl(\sum_{t} r(s_t^a,a_t^a)\Bigr)}{\exp\!\Bigl(\sum_{t} r(s_t^a,a_t^a)\Bigr) \;+\; \exp\!\Bigl(\sum_{t} r(s_t^b,a_t^b)\Bigr)},
  \label{eq:pref_model}
\end{equation}
where \(r(\cdot)\) denotes the unknown true reward function. Our goal is to learn an approximation \(r_\nu(\cdot)\) parameterized by \(\nu\) that aligns with these preference labels. We train \(r_\nu\) using the Bradley-Terry cross-entropy loss, which minimizes the difference between the predicted preference probabilities (using \(r_\nu\)) and the observed pairwise labels. Once \(r_\nu\) is learned, any standard RL algorithm (e.g., SAC \cite{haarnoja2018soft}) can optimize a policy \(\pi_\theta\) to maximize
\[
  \mathbb{E}_{\tau \sim \pi_\theta} \Bigl[\sum_{t} r_\nu(s_t,a_t)\Bigr].
\]

\subsection{Representation Challenges for VLM Preferences}

In preference-based RL, labels typically come from either human annotators or a fixed scripted teacher model ~\cite{lee2021pebble} ~\cite{park2022surf}. Recently, large VLM have also been employed to provide preference feedback in a more scalable manner. However, presenting only a single final image (or a sequence of frames) to the VLM may not capture the entire motion of a trajectory, making it difficult to reliably judge the agent’s behavior. We therefore propose representing the entire trajectory by \emph{sketching} it onto a single image, ensuring that the VLM’s preference labels better reflect the agent’s motion over time.

\subsection{Agent Alignment}
Even with richer representations, a learned reward can become \emph{misaligned} if we ignore the agent’s evolving policy. As \(\pi_\theta\) improves,  it is essential to account for the data generated by the current policy when shaping \(r_\nu\). In principle, one could formulate the process as a bilevel problem, early identified by ~\cite{chakraborty2023parl}:
\begin{equation}
\begin{aligned}
  \max_\nu \;\; & \mathbb{E}_{\tau \sim \pi_{\theta^*(\nu)}}\!\bigl[P(\tau^a \succ \tau^b)\bigr],\\
  \text{subject to}\;\; &
      \theta^*(\nu) \;=\; \arg\max_{\theta}\,\mathbb{E}_{\tau \sim \pi_\theta}[r_\nu(\tau)],
\end{aligned}
\label{eq:bilevel}
\end{equation}
but solving this exactly is computationally hard. Ignoring the agent’s performance during reward optimization can lead to suboptimal reward alignment, underscoring the need to incorporate agent preference in the learning process.

In subsequent sections, we will detail how these elements—pairwise preference modeling, augmented trajectory sketches, and iterative learning—come together to enable efficient, robust reward learning in robotic environments.

%% file: sections/5.system_overview.tex
\section{Method: VARP}
\label{sec:method}

We now present \emph{VARP}, an approach that addresses the practical challenges of preference-based RL in robotics by (1) incorporating a VLM with 2D trajectory sketches to improve feedback and (2) regularizing the learned reward to remain aligned with the agent’s evolving policy. This section explains how these components fit together into a unified framework.
\begin{figure}[t]
  \noindent
  \includegraphics[width=\columnwidth]{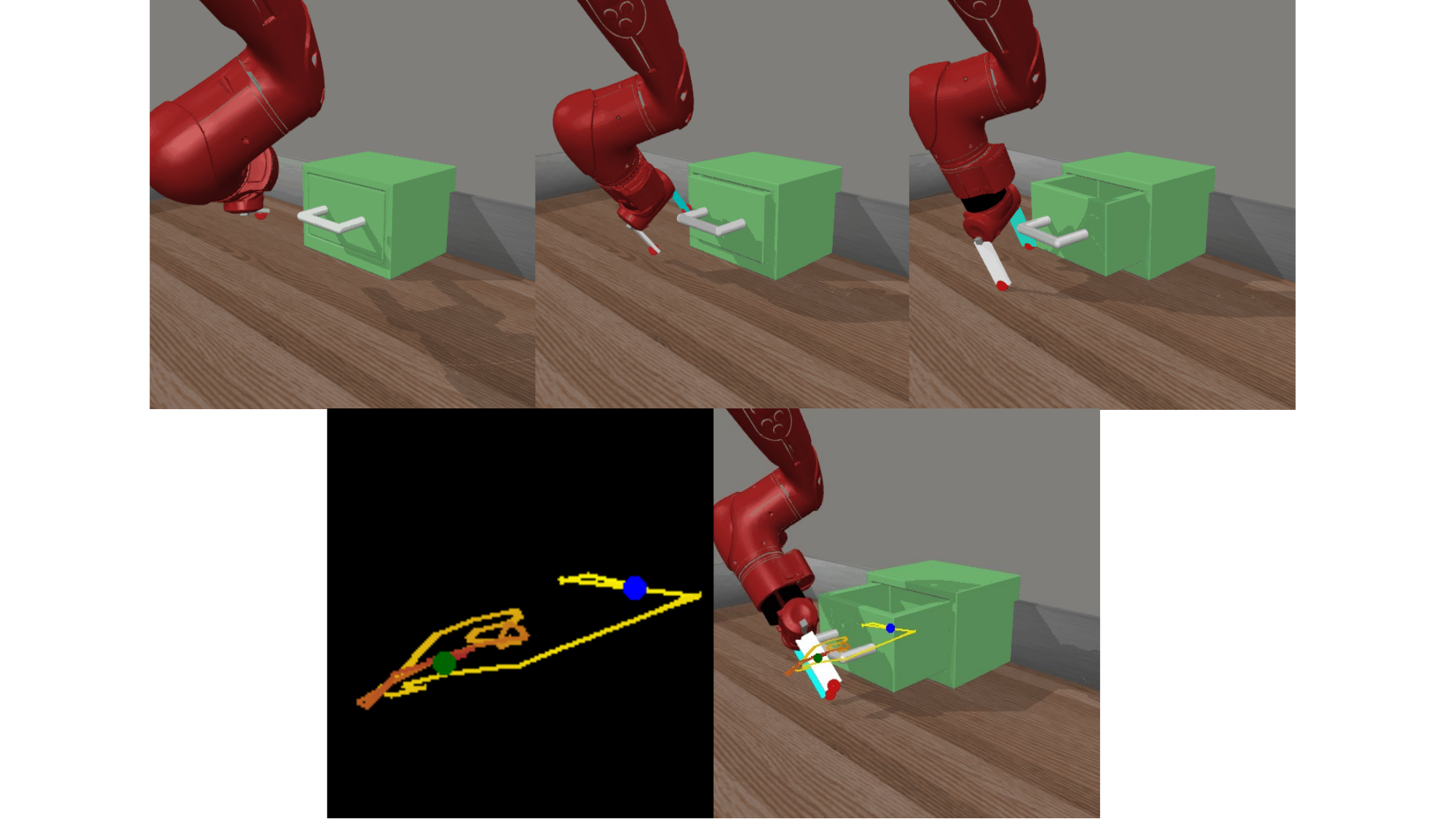}
  \caption{Illustration of 2D trajectory sketch generation. For each episode, the robot’s full trajectory (denoted by $\tau$) is projected onto the camera’s 2D plane using known parameters, and overlaid on the final state image $o$ to form an augmented observation $\hat{o}=(o,\mathrm{Sketch}(\tau))$. This enriched representation provides additional temporal context, enabling the VLM to more accurately compare and assess trajectory performance.}
  \label{fig:sketch_ex}
\end{figure}

% comment Syed Zaidi
% While the two-stage prompting improves interpretability, what happens if the VLM hallucinates a preference that contradicts actual task success? Maybe it should be tested In the cases where the VLM misinterprets ambiguous trajectories, and if so, how can inconsistencies in labeling be addressed ?

\subsection{Overview of the Pipeline}
Below, we outline the key steps of our pipeline, which are further elaborated in the following subsections and illustrated in Figure~\ref{fig:sketch_rl}.
\begin{enumerate}
\item \textbf{Generating 2D Trajectory Sketches:} Our setup involves a robot observed by an environment camera that captures image frames throughout each episode. As the policy $\pi_\theta$ runs, we record each transition $(s, a, s')$, storing relevant state information such as the end-effector's position or key object poses. This stream of data allows us to generate a 2D sketch, which is a trace of the end-effector’s path over time. We overlay this sketch on the \emph{final image} (i.e., the last camera frame of the episode) to provide a visual summary of the entire trajectory. 

\item \textbf{Preference Query via VLM:} Given two rollouts, we pair their final images (each annotated with the corresponding sketch) to form sketched observations $(\hat{o}^a, \hat{o}^b)$ \ref{fig:sketch_ex}. We then feed these into a VLM along with a text prompt describing the task objective.
Following RL-VLM-F \cite{wang2024rl}, we adopt a two-stage prompting strategy:
  \begin{itemize}
     \item In the \emph{analysis stage}, the VLM generates a free-form response describing how well each image meets the task goal.
     \item In the \emph{labeling stage}, we prompt the VLM again with the prior analysis to extract a single preference label $y \in \{-1, 0, 1\}$. 
  \end{itemize}
  A label of \(-1\) indicates no preference, and such pairs are discarded. For $y=0$ or $y=1$, we store the preference $(\tau^a,\tau^b,y)$ in a dataset $D$.
  % \item \textbf{Preference Query via VLM:} We pair each final image with the corresponding sketch, then feed pairs of these “sketched” observations $(\hat{o}^a, \hat{o}^b)$ into a VLM with a text prompt about the goal. The VLM outputs a preference label $y \in \{-1,0,1\}$, indicating which trajectory segment better achieves the goal.
  \item \textbf{Reward Model Training:} We learn a reward function $r_\nu$ that aligns with the VLM’s preference labels, while also incorporating \emph{agent preference} to avoid reward misalignment.
  \item \textbf{Policy Optimization:} The agent uses the newly updated reward to improve $\pi_\theta$, typically via an off-policy RL algorithm (SAC). The cycle repeats, collecting new data under the updated policy and refining $r_\nu$.
\end{enumerate}

By iterating these steps, VARP leverages VLM-based preferences \emph{and} trajectory sketches to generate more accurate feedback, while an agent preference term keeps the reward aligned with actual task success.

\subsection{Generating 2D Trajectory Sketches}

One of the key aspects of VARP is providing the VLM with richer trajectory representations that capture the agent’s entire motion, enabling more reliable preference judgments. At the end of each episode, we have a sequence of robot states and actions $\{(s_t,a_t)\}_{t=0}^H$ that define the episode’s trajectory $\tau$. We project the robot’s motion onto the camera’s 2D plane using known camera parameters (focal length, orientation, position). We then
 draw these coordinates on top of a selected environment
 frame, color-coding the connecting lines so that their hue
 shifts from bright yellow at the start to a darker tone at
 the end, reflecting temporal progression. A simpler alternative is to track pixel coorinates of one of the end-effector, but that can lead to some inaccurate sketches. Initial and final position are optionally marked with colored circles, one of the examples is shown in Figure~\ref{fig:sketch_ex}. Figure~\ref{fig:sketch_rl} (left) illustrates a scenario in which two final states appear nearly identical. Adding sketches to the same final images highlights the path taken, clarifying which trajectory made more efficient or successful progress. We hypothesize that visually depicting \emph{how} each final state was reached reduces ambiguity in partially completed tasks and near-identical outcomes, enabling more precise VLM comparisons.

%  Formally:
% \[
%   \hat{o} = \bigl(o,\;\mathrm{Sketch}(\tau)\bigr),
% \]
% where $o$ is the raw final image and $\mathrm{Sketch}(\tau)$ is the overlay function. Presenting $(\hat{o}^a, \hat{o}^b)$ to the VLM exposes \emph{how} each final state was achieved, reducing ambiguity in partial completions or near-identical end states.

\subsection{Reward Regularization}

\begin{algorithm}[t]
\caption{VARP}
\label{alg:sketch-rl-varp}
\textbf{Inputs:} Text description $\ell$, regularization $\lambda$, initial policy $\pi_\theta$, reward model $r_\nu$, replay buffer $B$, preference dataset $D=\emptyset$, number of iterations $T$.

\begin{algorithmic}[1]
\FOR{$\text{iteration} \;=\; 1, 2, \dots, T$}
    \STATE \textbf{Sketch Data Generation:}
    \begin{enumerate}
        \item Roll out $\pi_\theta$ in the environment to gather transitions $(s,a,s')$ for multiple episodes.
        \item For each episode, record the final image $o$ and generate a \emph{2D sketch} of the trajectory $\tau$ to form $\hat{o} = \bigl(o,\;\mathrm{Sketch}(\tau)\bigr)$.
        \item Store $(s,a,s'), \hat{o}$ in the replay buffer $B$.
    \end{enumerate}

    \STATE \textbf{VLM Preference Query (2-Stage):}
    \begin{enumerate}
        \item Sample pairs $(\hat{o}^a, \hat{o}^b)$ from $B$.
        \item \emph{Analysis Stage:} Prompt the VLM to describe and compare each image+sketch relative to the goal.
        \item \emph{Labeling Stage:} Prompt the VLM again with its analysis to obtain a preference $y \in \{-1,0,1\}$.
        \item If $y = -1$ (no difference), discard; else store $(\tau^a,\tau^b,y)$ in $D$.
    \end{enumerate}

    \STATE \textbf{Reward Update:}
    \begin{enumerate}
      \item Compute $\mathcal{L}_{\text{VLM}}(\nu)$ per \eqref{eq:vlm_loss}
      \item Compute $\mathcal{L}_{\text{agent}}(\nu)$ per \eqref{eq:agent_loss}
      \item Compute $\mathcal{L}_{\text{VARP}}(\nu)$ per \eqref{eq:varp_loss}
      \item Update $r_\nu$ by minimizing $\mathcal{L}_{\text{VARP}}(\nu)$
    \end{enumerate}

    \STATE \textbf{Policy Update:}
    \begin{enumerate}
        \item Relabel all transitions in $B$ using the updated reward $r_\nu(s,a)$.
        \item Optimize $\pi_\theta$ (SAC) to maximize $\mathbb{E}\bigl[\sum_t r_\nu(s_t,a_t)\bigr]$.
    \end{enumerate}
\ENDFOR

\RETURN $\pi_\theta, \; r_\nu$
\end{algorithmic}
\end{algorithm}

\noindent
\textbf{Bilevel Optimization Framework.}
We train two distinct models in a nested (two-loop) way:
\begin{enumerate}
  \item \textbf{Reward Model} $r_\nu$: Updated in the \emph{outer loop} by incorporating both VLM-derived preferences and agent constraints.
  \item \textbf{Policy} $\pi_\theta$: Optimized in the \emph{inner loop} to maximize the learned reward.
\end{enumerate}

\paragraph*{Outer Loop (Reward Model Update).}
Let $D = \{(\tau^a, \tau^b, y)\}$ be a dataset of trajectory pairs and preference labels (VLM). We define:
\begin{equation}
\begin{aligned}
  \mathcal{L}_{\text{VLM}}(\nu) 
  \;=\;
  -\sum_{(\tau^a,\tau^b,y)\in D}
    \Bigl[
    &\,y\;\log P_\nu(\tau^a \succ \tau^b)\\
    &\quad+\;
    (1-y)\;\log P_\nu(\tau^b \succ \tau^a)
    \Bigr].
\end{aligned}
\label{eq:vlm_loss}
\end{equation}
where $P_\nu(\tau^a \succ \tau^b)$ is derived from a Bradley-Terry model using the cumulative reward from $r_\nu$ \cite{chakraborty2024parl}. This encourages

\begin{figure*}[t!]
\centering

\begin{subfigure}{0.7\textwidth}
    \includegraphics[width=\linewidth]{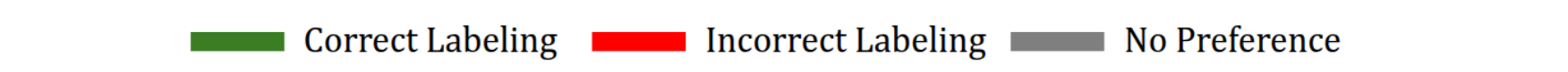}
\end{subfigure} 
\vspace{1em}

% Top row
\begin{subfigure}[b]{0.32\textwidth}
  \centering
  \includegraphics[width=\linewidth]{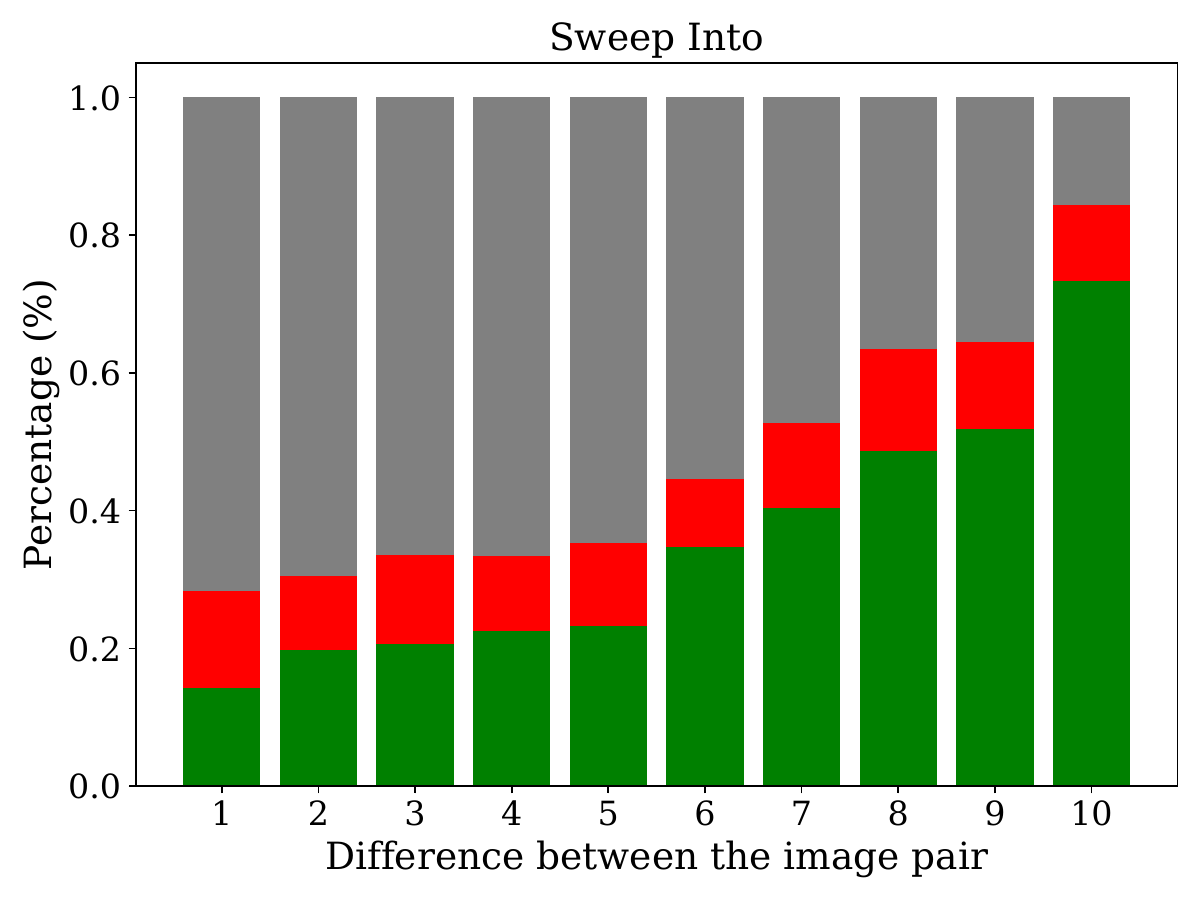}
\end{subfigure}
\hfill
\begin{subfigure}[b]{0.32\textwidth}
  \centering
  \includegraphics[width=\linewidth]{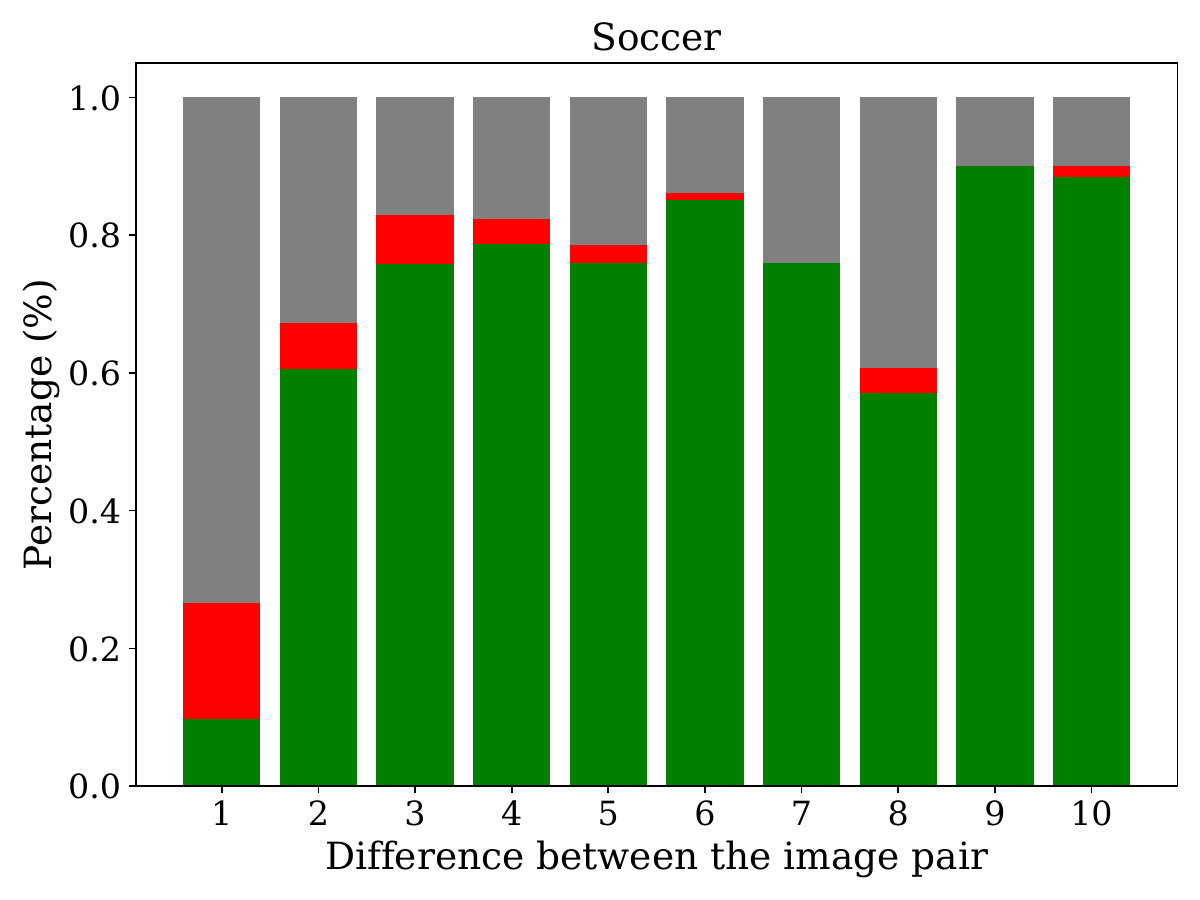}
\end{subfigure}
\hfill
\begin{subfigure}[b]{0.32\textwidth}
  \centering
  \includegraphics[width=\linewidth]{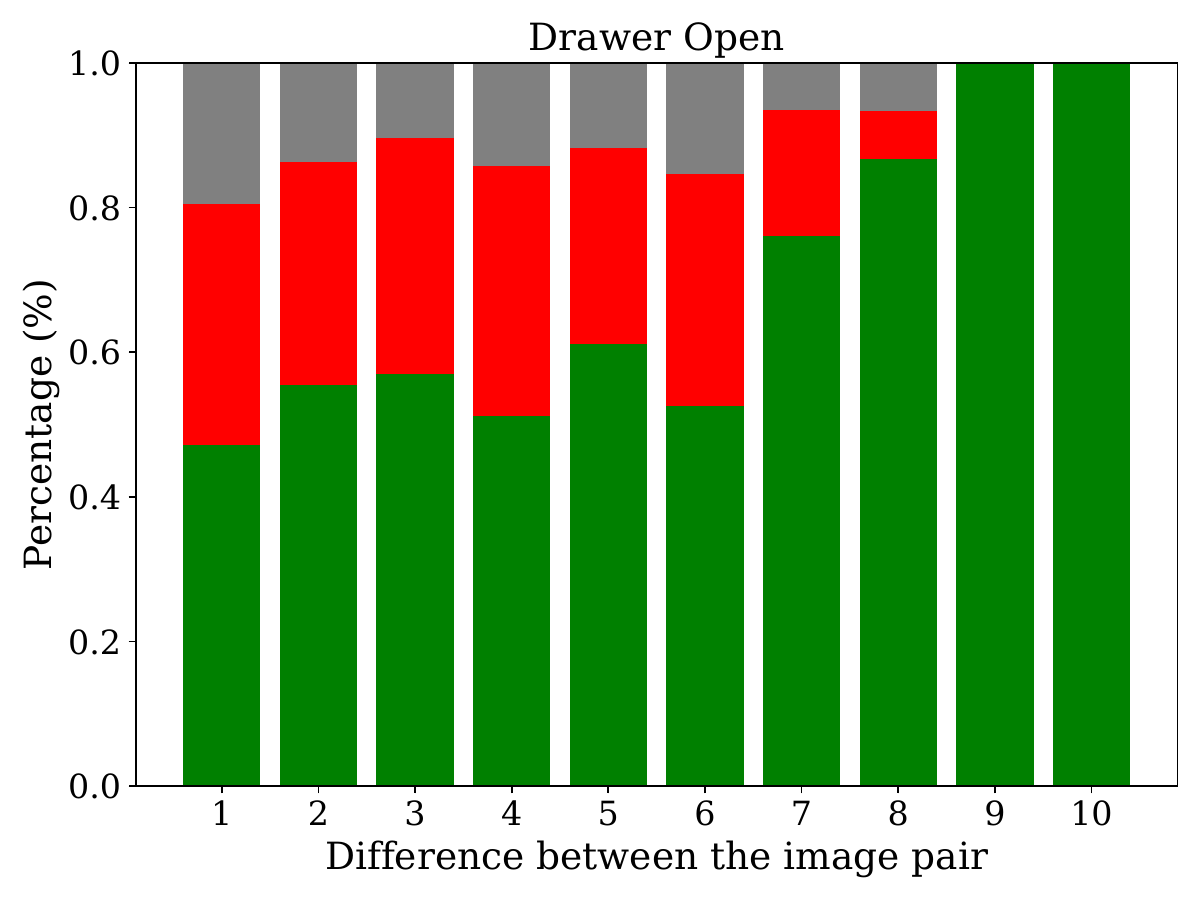}
\end{subfigure}

\vspace{1em}

% Bottom row
\begin{subfigure}[b]{0.32\textwidth}
  \centering
  \includegraphics[width=\linewidth]{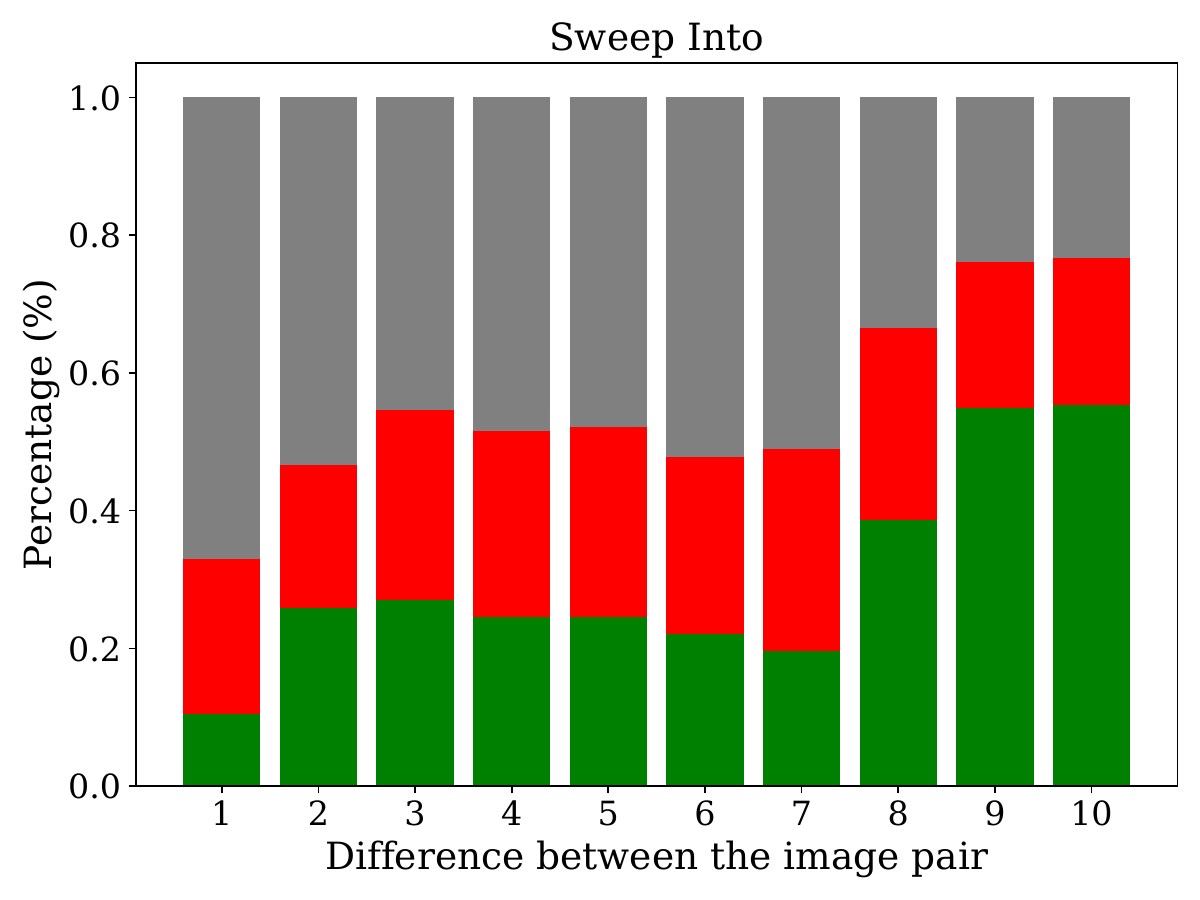}
\end{subfigure}
\hfill
\begin{subfigure}[b]{0.32\textwidth}
  \centering
  \includegraphics[width=\linewidth]{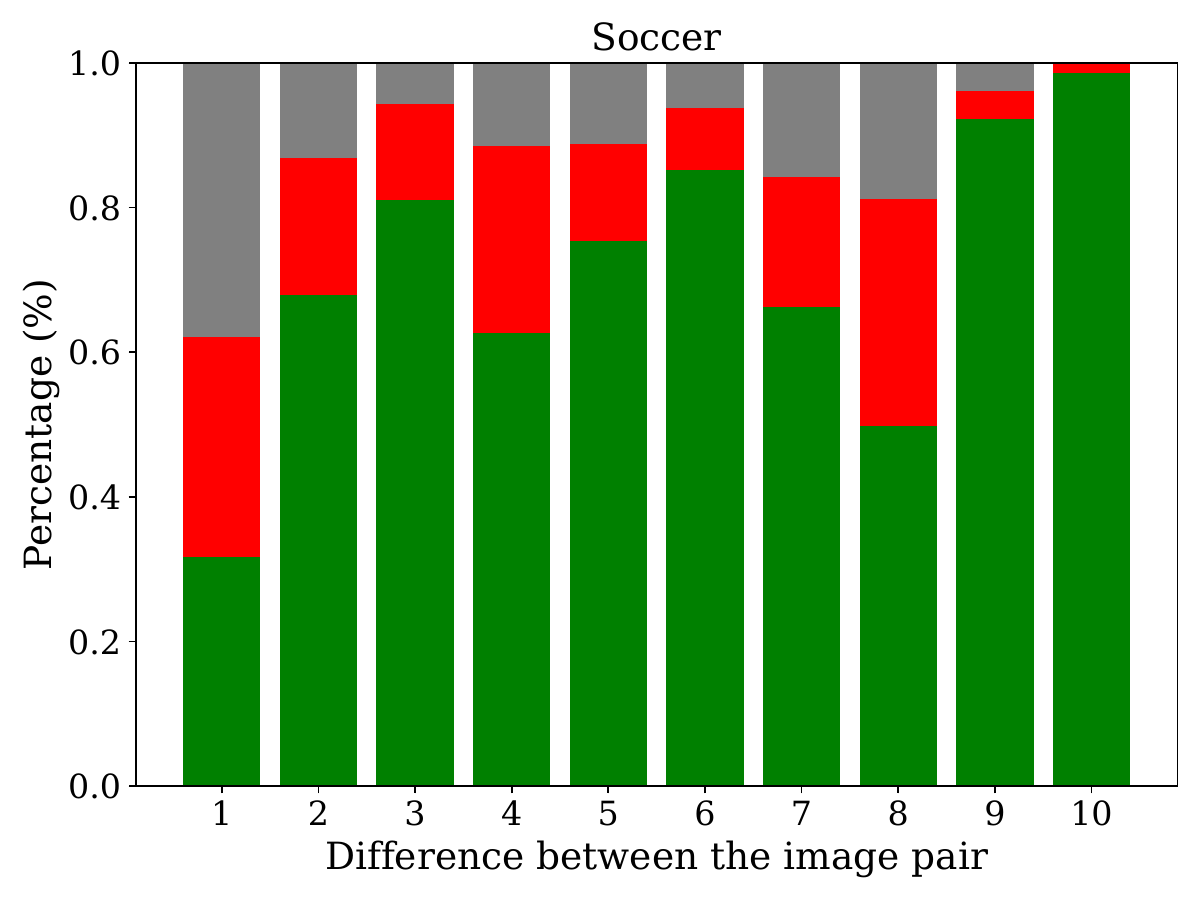}
\end{subfigure}
\hfill
\begin{subfigure}[b]{0.32\textwidth}
  \centering
  \includegraphics[width=\linewidth]{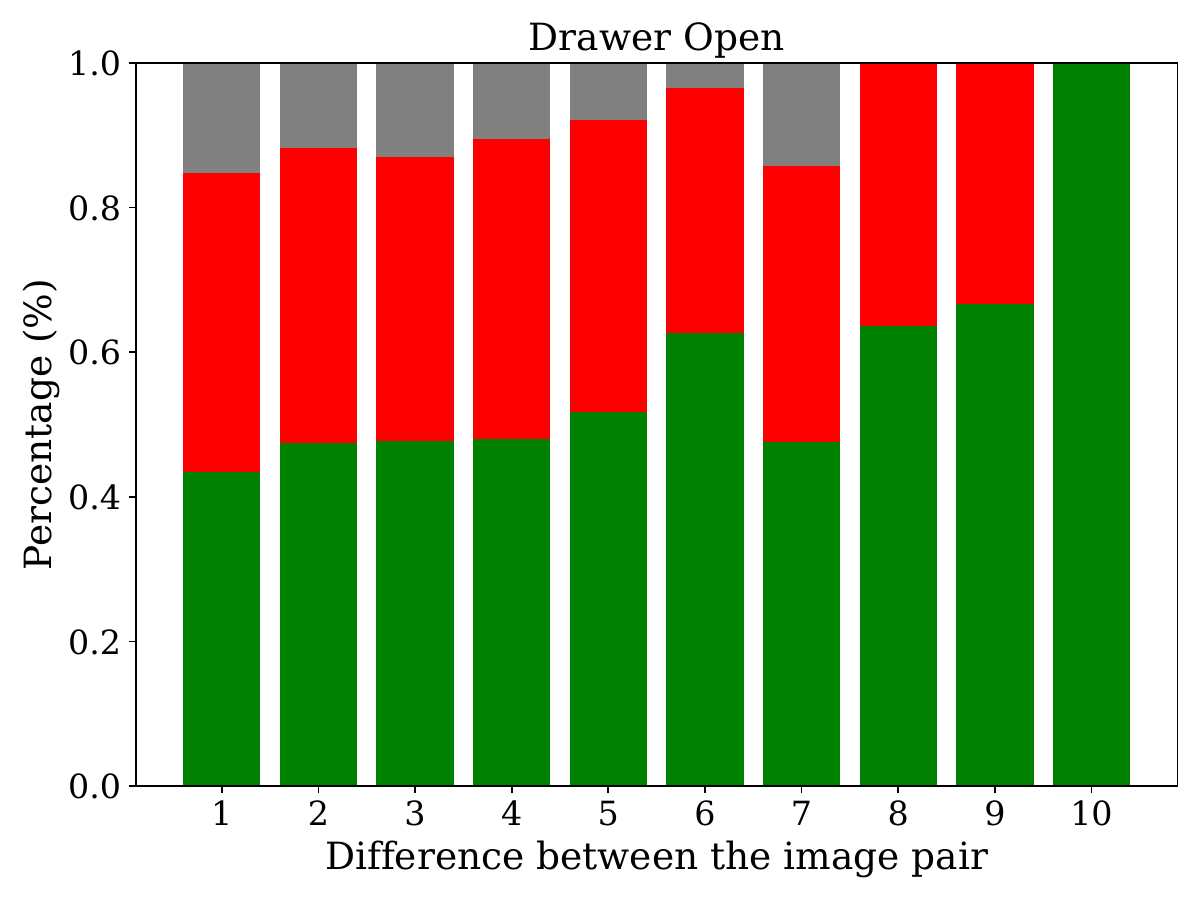}
\end{subfigure}

\caption{\textbf{Impact of Trajectory Sketches on Preference Accuracy.}
Top row: VLM predictions using trajectory sketches. Bottom row: VLM predictions without sketches. The results show that incorporating sketches dramatically improves the accuracy of preference judgments, particularly when the difference in task progress between image pairs increases. This confirms that visualizing the entire trajectory—not just the final state—provides essential context for reliable feedback.}
\label{fig:pref_curves}
\end{figure*}

the reward to rank $\tau^a$ above $\tau^b$ whenever $y=1$, and vice versa.

Next, we introduce an \emph{agent preference} regularizer to maintain alignment with the agent’s achievable behaviors:
\begin{equation}
\begin{aligned}
  \mathcal{L}_{\text{agent}}(\nu) 
  &= \lambda 
    \,\mathbb{E}_{\tau \sim \pi_\theta}
      \bigl[-R(\tau)\bigr], \\
  \text{where } 
  R(\tau) 
  &= \sum_{t}r_\nu(s_t, a_t).
\end{aligned}
\label{eq:agent_loss}
\end{equation}
The hyperparameter $\lambda$ weighs the importance of not overvaluing low-return trajectories under $\pi_\theta$. Concretely, the second term grows when the agent’s return is small, preventing $r_\nu$ from assigning large rewards to unsuccessful behaviors and keeping the reward function focused on truly effective trajectories. We combine these terms into a single reward-model loss:
\begin{equation}
  \mathcal{L}_{\text{VARP}}(\nu) 
  \;=\;
    \mathcal{L}_{\text{VLM}}(\nu)
    \;+\;
    \mathcal{L}_{\text{agent}}(\nu).
  \label{eq:varp_loss}
\end{equation}
Minimizing $\mathcal{L}{\text{VARP}}(\nu)$ with respect to $\nu$ results in an updated reward $r\nu$ that balances external preferences with viable agent behaviors.

\paragraph*{Inner Loop (Policy Update).}
Holding the reward model $r_\nu$ fixed, we train the policy $\pi_\theta$ to maximize the expected return:
\begin{equation}
  \theta^*
  = \arg\max_{\theta}
    \mathbb{E}_{\tau \sim \pi_\theta}\Bigl[\,
      \sum_{t}r_\nu(s_t, a_t)
    \Bigr].
  \label{eq:policy_opt}
\end{equation}
This policy optimization is done using a standard RL algorithm, SAC ~\cite{haarnoja2018soft} in our case. By alternating between these two loops, the reward model remains anchored to both external feedback and agent performance, ensuring stable convergence and improved task alignment.

% Algorithm~\ref{alg:sketch-rl-varp} outlines the bilevel procedure:
% \begin{enumerate}
%   \item \emph{Outer Loop:} Update $\nu$ by minimizing $\mathcal{L}_{\text{VARP}}(\nu)$ in Eq.~\eqref{eq:varp_loss}.
%   \item \emph{Inner Loop:} Update $\theta$ by maximizing expected returns under the fixed $r_\nu$, as in Eq.~\eqref{eq:policy_opt}.
% \end{enumerate}
By alternating between these two loops, the reward model remains anchored to both external feedback and agent performance, ensuring stable convergence and improved task alignment.

%% file: sections/6.experiments.tex
\section{Experiments}
We structured our experiments around three core questions: 
\begin{enumerate} 
\item Does representing trajectories with sketches improve the accuracy of VLM-based preferences? 
\item Does incorporating sketches with agent preference also lead to better policy learning? 
\item Does adding agent preference help regularize reward learning in robotic Preference RL?
% to VLM-based preference learning help mitigate distribution shift and achieve more robust task alignment? 
\end{enumerate}

As an oracle reference, we also include ground-truth (GT) preference labels derived from each environment’s built-in reward function, serving as an upper bound. 
% Detailed prompts for the two-stage labeling process can be found in the Appendix.

\subsection{Experimental Setup}
% To investigate these questions, we test on a variety of manipulation tasks from MetaWorld~\cite{yu2020meta} and locomotion tasks from DMControl~\cite{tassa2018deepmind}. We compare our method, VARP and reward regularization formulation, against two main baselines:
% For the VLM-based methods, we use ChatGPT~\cite{openai2023chatgpt} and conduct experiments on three Meta-world manipulation tasks:
% \\
% \textit{Drawer Open:} The robot arm must reach the drawer handle and pull it to open.\\
% \textit{Soccer:} The robot arm needs to push a soccer ball into a designated goal.\\
% \textit{Sweep Into:} The robot arm must reach a green cube and sweep it into a hole on a wooden table.\\ 
% We also evaluate reward regularization formulation in DMControl locomotion tasks:\\
% \textit{Walker:} The reward function incentivizes forward movement of the robot.\\
% \textit{Cheetah:} The agent's objective is to achieve running speeds up to a specified threshold.\\
To investigate these questions, we conduct evaluations on a range of manipulation tasks from MetaWorld~\cite{yu2020meta} and locomotion tasks from DMControl~\cite{tassa2018deepmind}. We compare our method, VARP, and the reward regularization formulation against two main baselines.

\begin{figure*}[t!]
\centering
% Top row: Images 1, 2, 3
\begin{subfigure}{0.7\textwidth}
    \includegraphics[width=\linewidth]{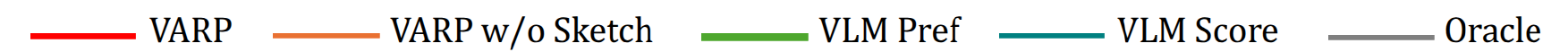}
\end{subfigure} 
\begin{subfigure}[b]{0.32\textwidth}
  \centering
  \includegraphics[width=\linewidth]{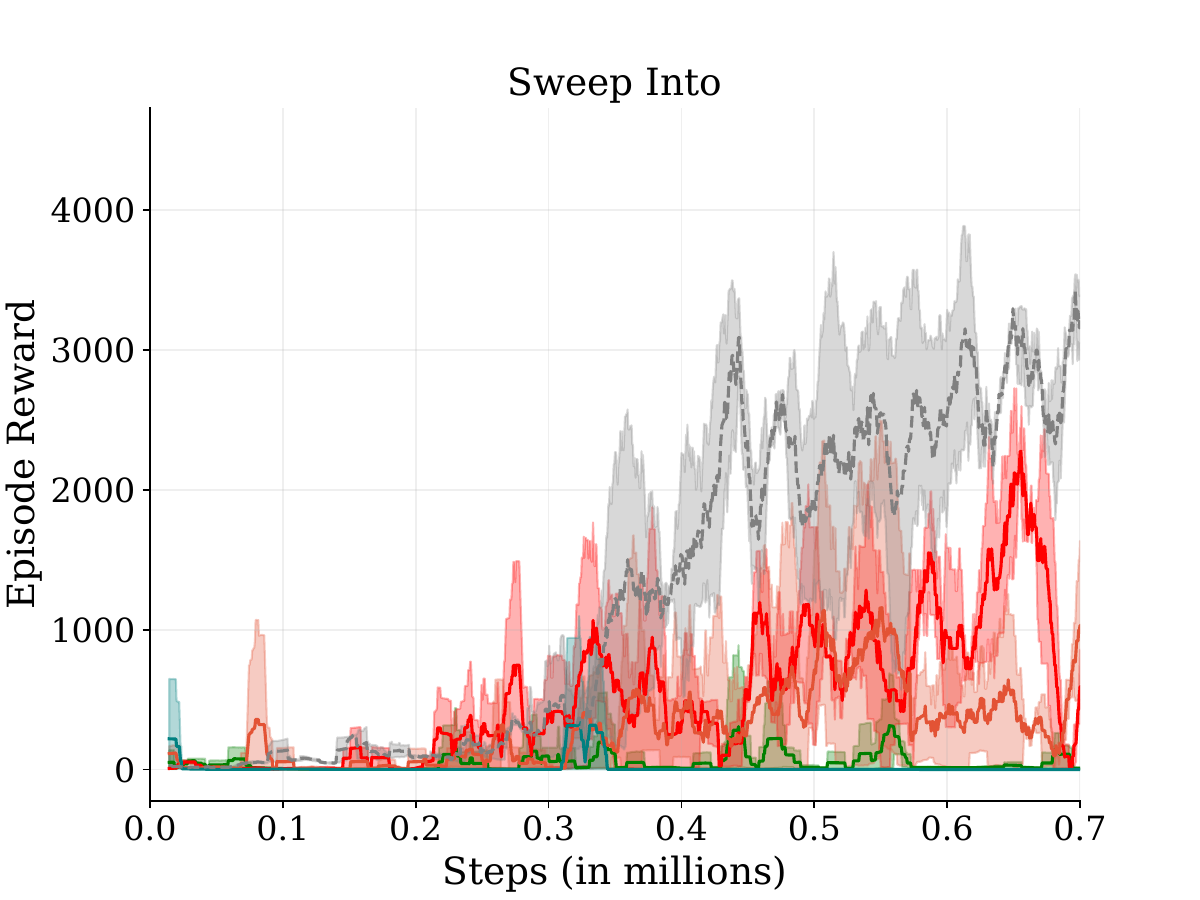}
  %\caption{Image 1}
\end{subfigure}
\hfill
\begin{subfigure}[b]{0.32\textwidth}
  \centering
  \includegraphics[width=\linewidth]{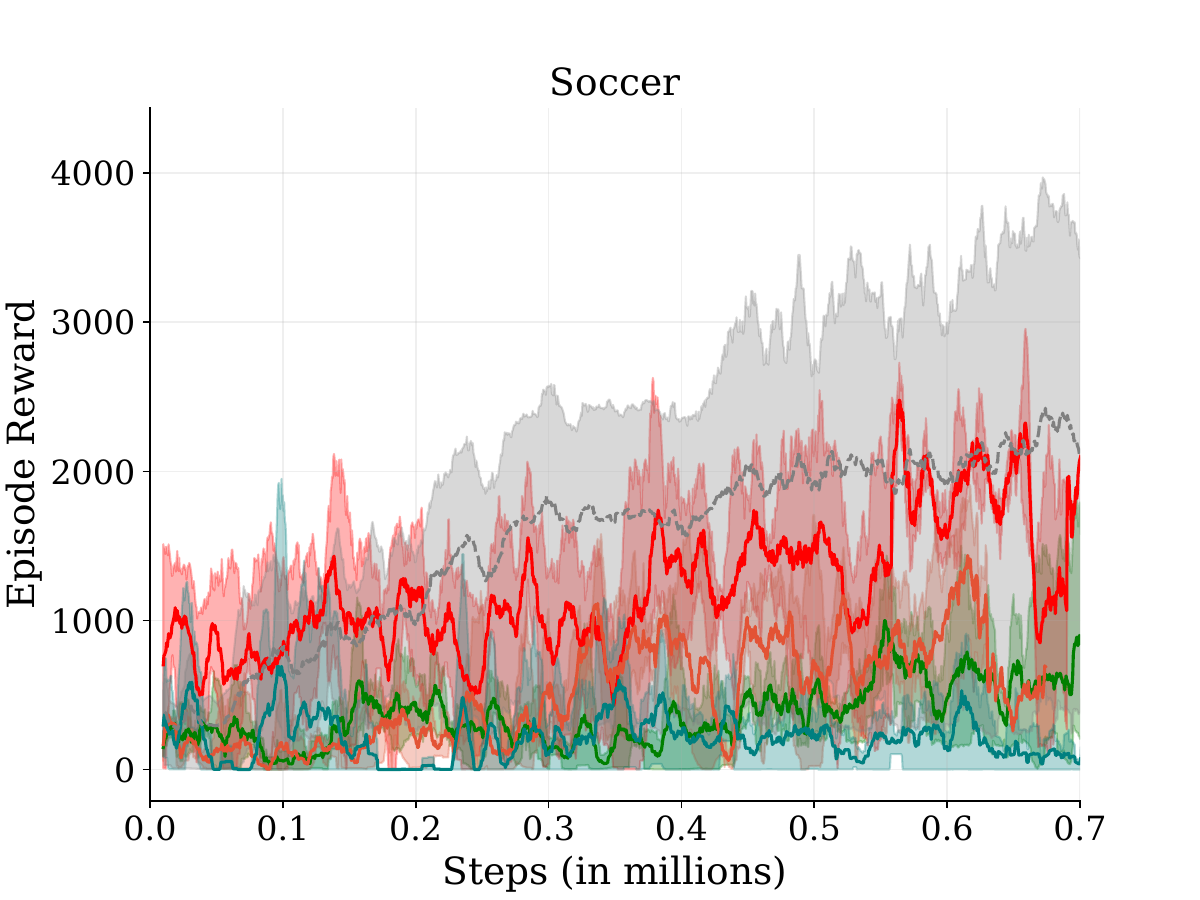}
  %\caption{Image 2}
\end{subfigure}
\hfill
\begin{subfigure}[b]{0.32\textwidth}
  \centering
  \includegraphics[width=\linewidth]{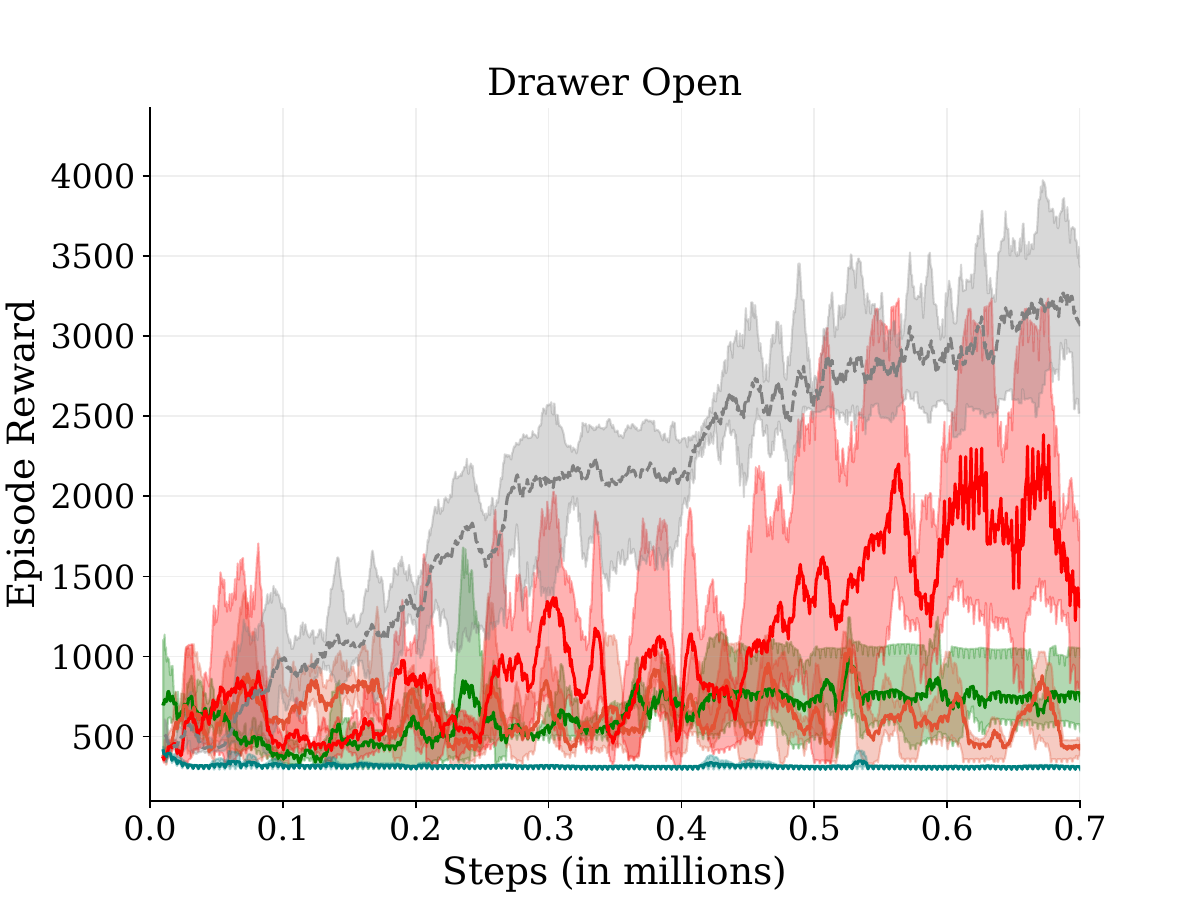}
  %\caption{Image 3}
\end{subfigure}

\vspace{1em} % Adjust vertical spacing between rows if needed

% Bottom row: Images 4, 5, 6
\begin{subfigure}[b]{0.32\textwidth}
  \centering
  \includegraphics[width=\linewidth]{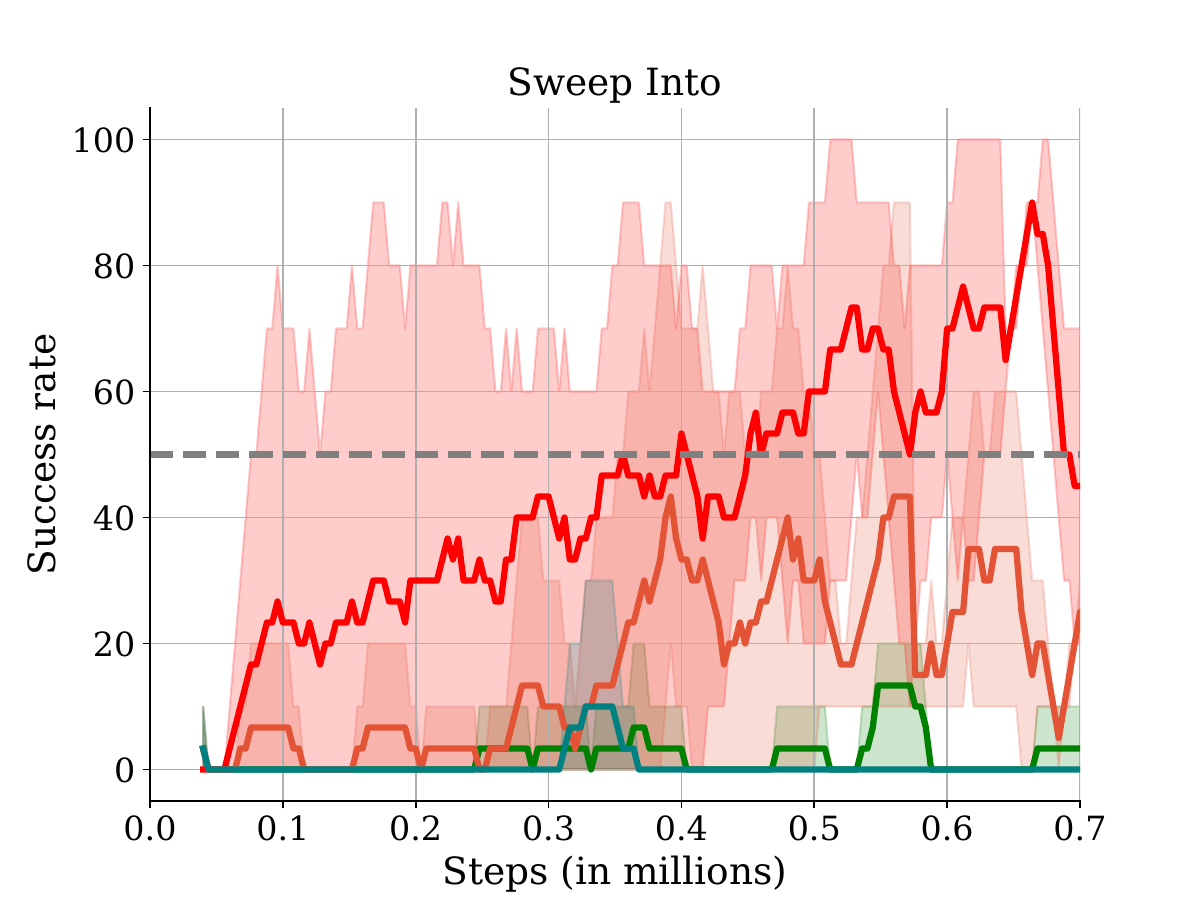}
  %\caption{Image 4}
\end{subfigure}
\hfill
\begin{subfigure}[b]{0.32\textwidth}
  \centering
  \includegraphics[width=\linewidth]{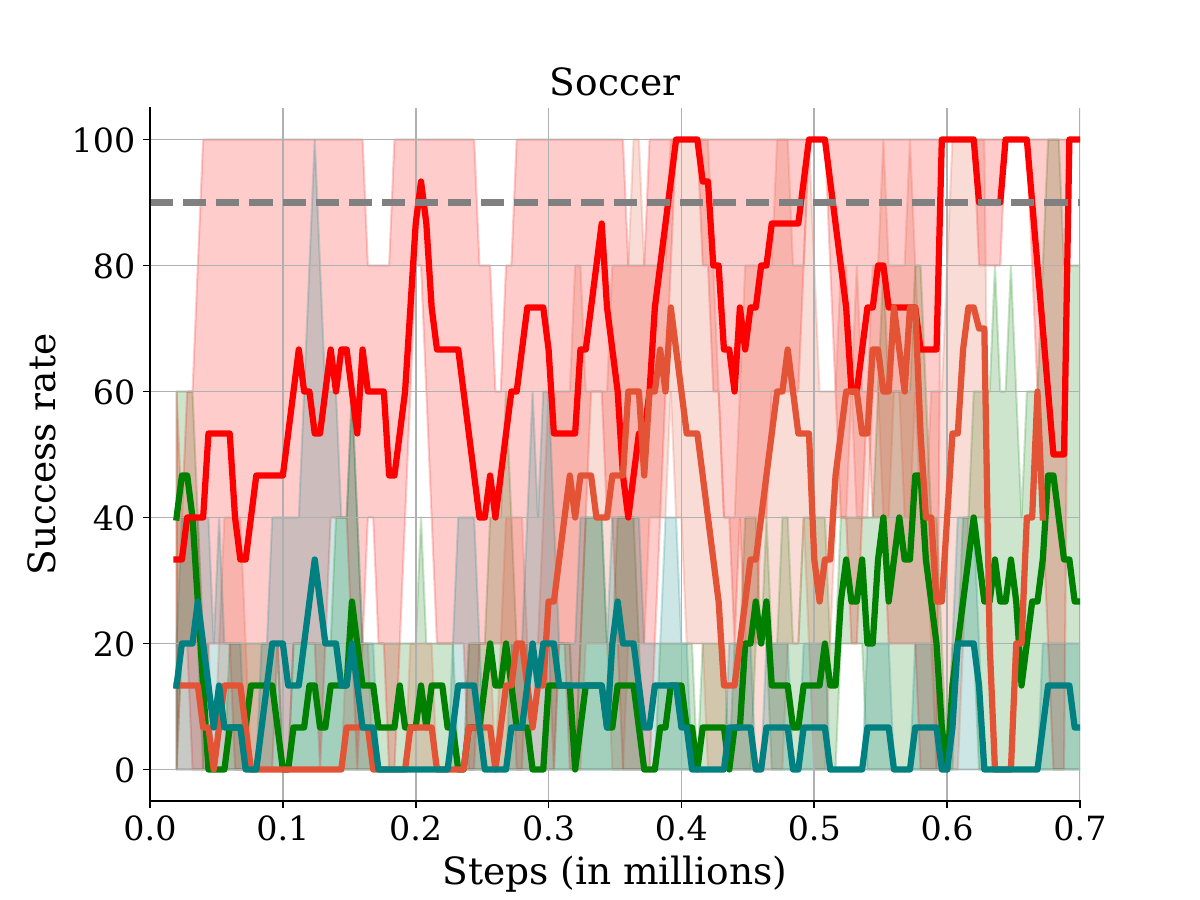}
  %\caption{Image 5}
\end{subfigure}
\hfill
\begin{subfigure}[b]{0.32\textwidth}
  \centering
  \includegraphics[width=\linewidth]{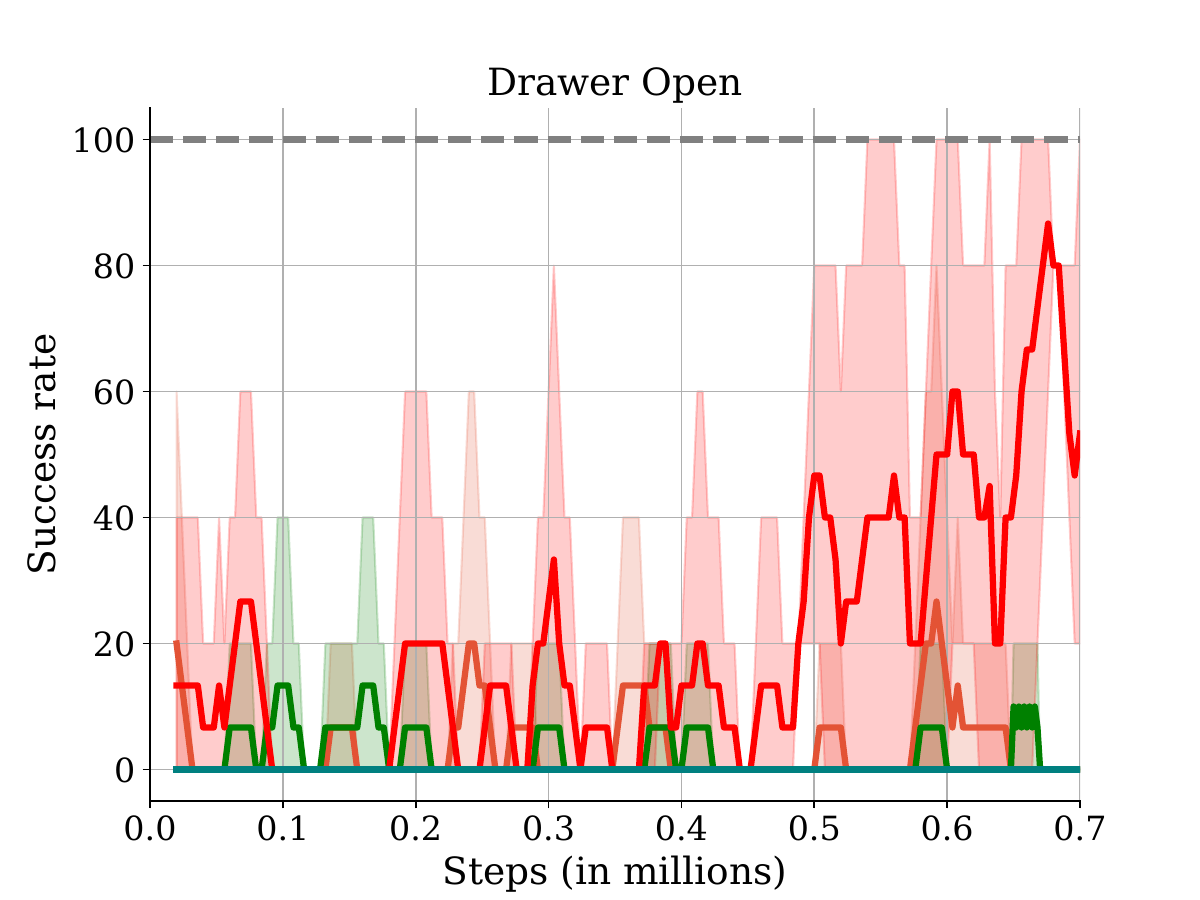}
  %\caption{Image 6}
\end{subfigure}

\caption{ \textbf{Performance Comparison on MetaWorld Tasks.}
We compare VARP against baseline approaches across three MetaWorld tasks (Drawer Open, Soccer, Sweep Into). Our method, which combines trajectory sketches with agent preference regularization, consistently achieves higher episode rewards. The plot highlights how the enhanced preference accuracy translates into faster training and better overall policy performance, closing the gap to oracle-level behavior. }
\label{fig:training_curves}
\end{figure*}

For VLM-based methods, we use ChatGPT~\cite{openai2023chatgpt} model 4o and perform experiments on three MetaWorld manipulation tasks:

\begin{itemize}
    \item \textbf{Drawer Open:} The robot arm must reach the drawer handle and pull it to open.
    \item \textbf{Soccer:} The robot arm needs to push a soccer ball into a designated goal.
    \item \textbf{Sweep Into:} The robot arm must reach a green cube and sweep it into a hole on a wooden table.
\end{itemize}

Additionally, we evaluate the reward regularization formulation on locomotion tasks from DMControl:

\begin{itemize}
    \item \textbf{Walker:} The reward function incentivizes forward movement of the robot.
    \item \textbf{Cheetah:} The agent's objective is to achieve running speeds up to a specified threshold.
\end{itemize}

Also, to note, in the manipulation task, the start position of the arm is random and is also not restricted to always being in the view. As an oracle benchmark, we use ground truth preferences derived from each environment's reward function. These preferences serve as an upper bound to evaluate our method's performance.

\subsection{Does representing trajectories with sketches improve the accuracy of VLM-based preferences?}
In this experiment, we demonstrate how augmenting final image observations with sketches provides crucial information for preference learning. We evaluate this by comparing VLM preferences with and without trajectory sketches in our pipeline (Figure~\ref{fig:pref_curves}). When the difference between a pair of images (measured by the difference between the rewards of to trajectories) is minimal, both human and VLM annotators struggle to provide accurate preferences based on final states alone. Our results indicate that using final images alone yields an average preference accuracy of 68\% for near-identical end states. In contrast, incorporating sketched trajectories increases this accuracy to 84\%. Note that if the VLM indicates “no preference” ($y=-1$), we discard the pair.
The plot in Figure \ref{fig:pref_curves} clearly demonstrates that incorporating trajectory sketches reduces instances of incorrect preferences across all scenarios. However, their benefit is particularly pronounced when rewards of two trajectories are similar (bin 1 in Figure \ref{fig:pref_curves}). As expected, when the difference between trajectories is substantial, sketches further enhance the VLM's ability to provide correct preferences by making the distinctions even more apparent. This confirms our hypothesis that visualizing the full trajectory, rather than just the final state, provides essential context for more accurate preference judgments.

In Figure \ref{alg:sketch-rl-varp} we show an example of the VLM response and preferences given a pair of images with and without sketches. Due to the additional modality the VLM is able to provide a better preference in terms of safety and efficiency.

% In this experiment, we demonstrate how augmenting final image observations with sketches provides crucial information for preference learning. We evaluate this by comparing VLM preferences with and without trajectory sketches in our pipeline (Figure~\ref{fig:pref_curves}).

% When the difference between a pair of images (as measured by task progress) is minimal, both human and VLM annotators struggle to provide accurate preferences based on final states alone. Our results indicate that using final images alone yields an average preference accuracy of 68% for near-identical end states. In contrast, incorporating sketched trajectories increases this accuracy to 84%—a relative improvement of 16%. Note that if the VLM indicates “no preference” ($y=-1$), we discard the pair.
% The plot in Figure~\ref{fig:pref_curves} clearly demonstrates that incorporating trajectory sketches reduces instances of incorrect preferences across all scenarios. When the difference between trajectories is substantial, sketches further enhance the VLM's ability to provide correct preferences by making distinctions more apparent. This confirms our hypothesis that visualizing the full trajectory, rather than just the final state, provides essential context for more accurate preference judgments.

% In Figure~\ref{alg:sketch-rl-varp} we show an example of the VLM response and preferences given a pair of images with and without sketches.

\subsection{Does incorporating sketches with agent preference also lead to better policy learning?}

We next investigate whether the improved preference labels provided by VARP lead to higher-performing policies. To gauge performance, we use two metrics: \emph{episode reward} (the sum of the learned reward over an episode) and

\begin{figure*}[t!]
\centering
\begin{subfigure}{0.7\textwidth}
    \includegraphics[width=\linewidth]{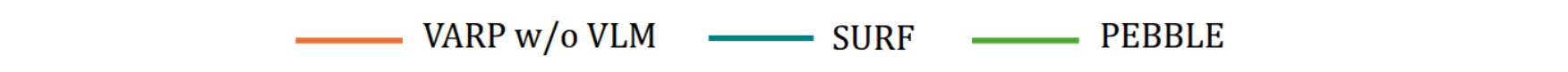}
\end{subfigure} 
% Top row: Images 1, 2, 3
\begin{subfigure}[b]{0.32\textwidth}
  \centering
  \includegraphics[width=\linewidth]{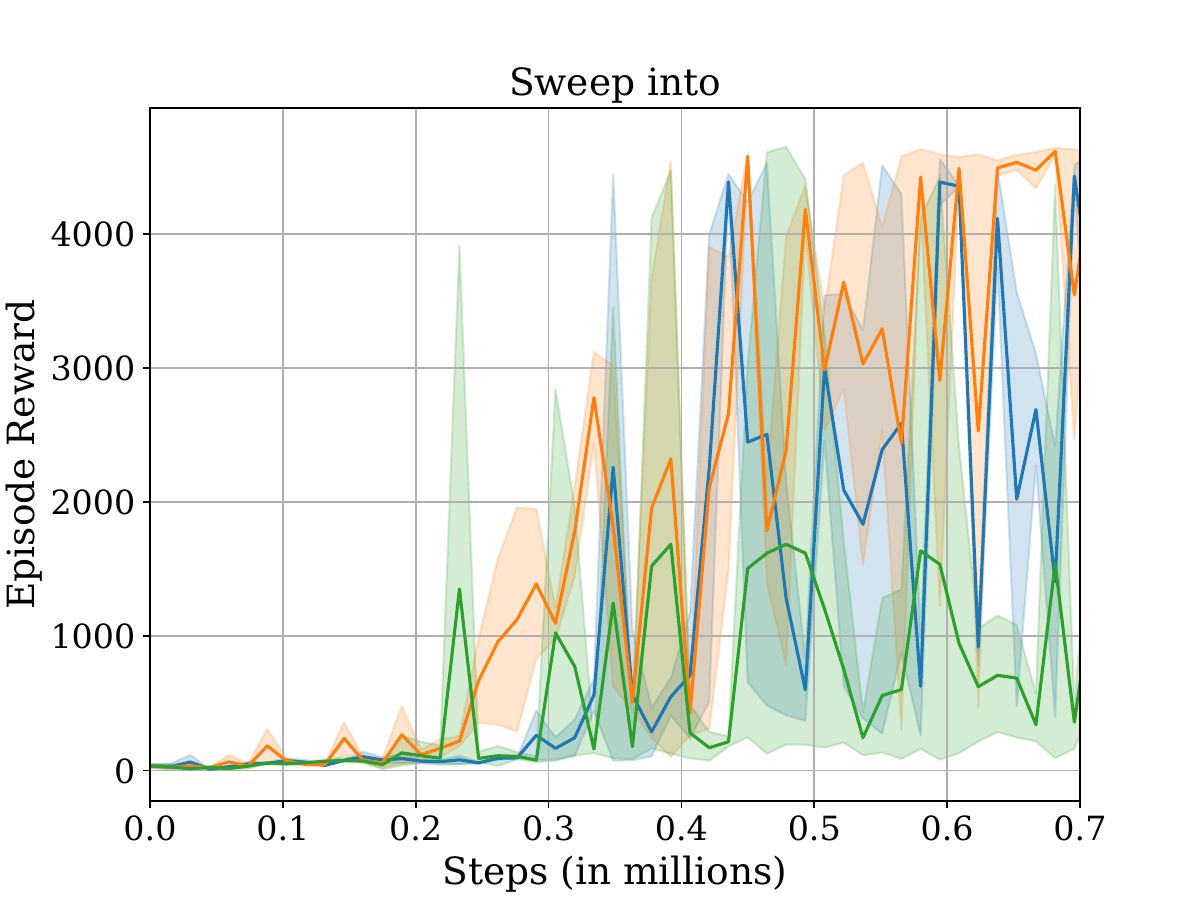}
  %\caption{Image 1}
\end{subfigure}
\hfill
\begin{subfigure}[b]{0.32\textwidth}
  \centering
  \includegraphics[width=\linewidth]{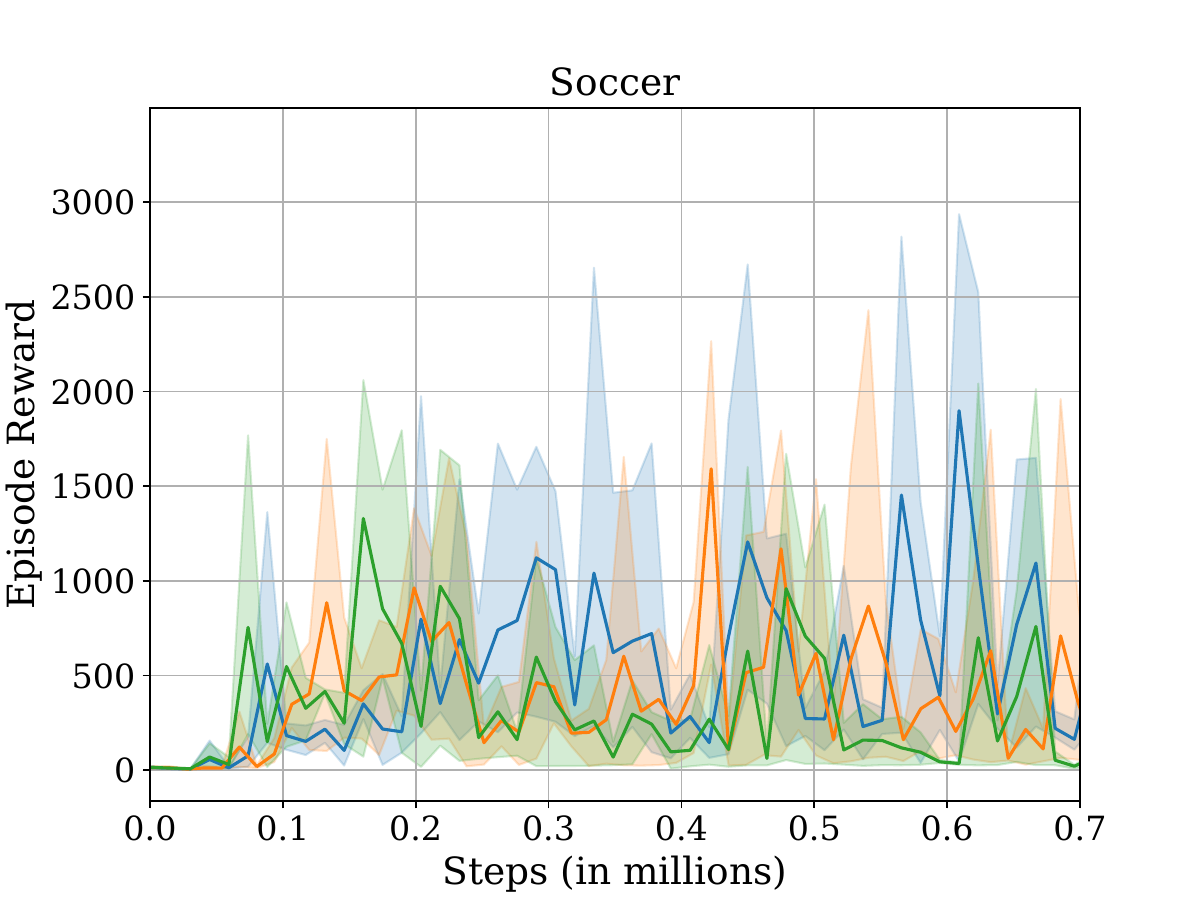}
  %\caption{Image 2}
\end{subfigure}
\hfill
\begin{subfigure}[b]{0.32\textwidth}
  \centering
  \includegraphics[width=\linewidth]{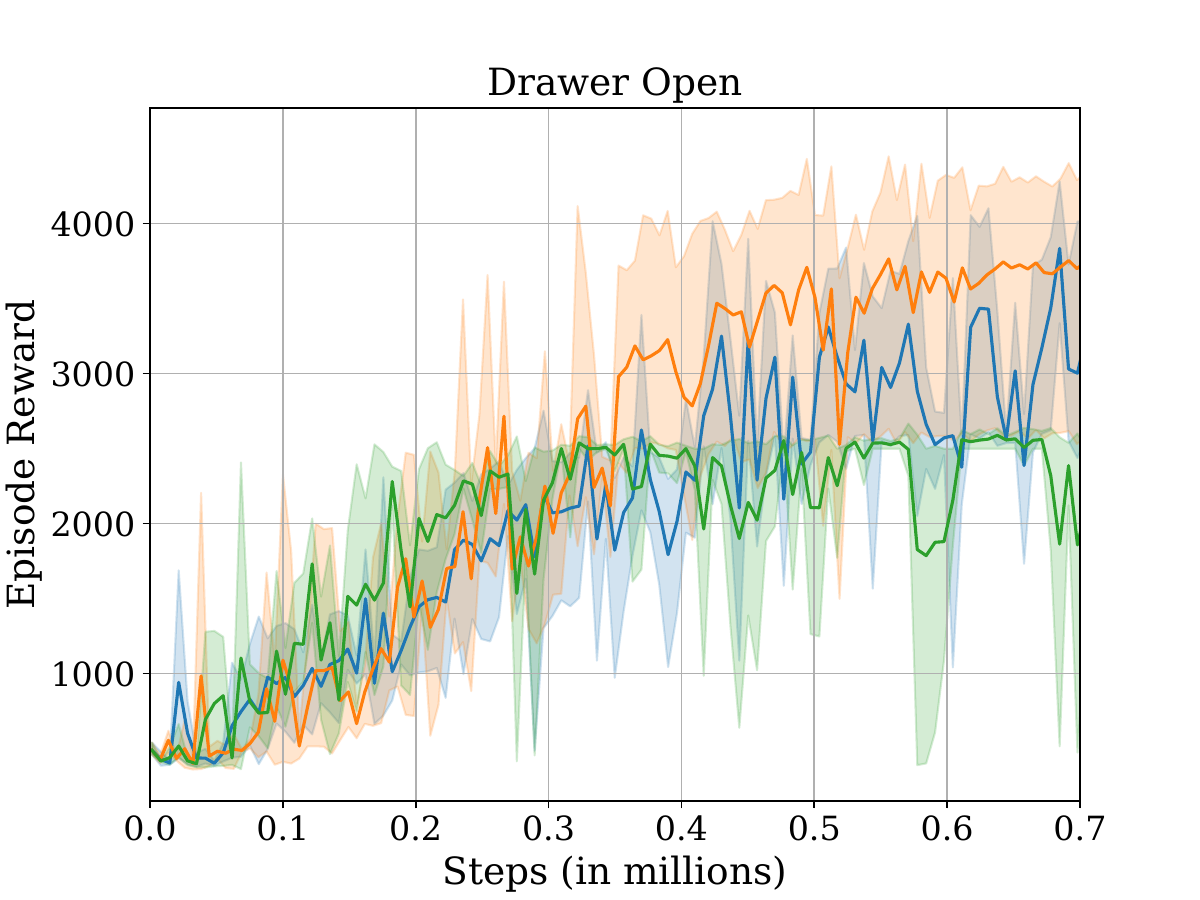}
  %\caption{Image 3}
\end{subfigure}
\caption{\textbf{Evaluating RLHF Accuracy with VARP in Metaworld.}
This figure quantifies the accuracy of our reward learning by comparing RLHF predictions to ground-truth preferences derived from the environment’s reward function. The results demonstrate that the agent preference substantially reduces reward misalignment (and hence reward hacking), ensuring more stable and effective policy improvements.}
\label{fig:rlhf_curves}
\end{figure*}

\emph{success rate} (based on task-specific completion criteria). As shown in Figure~\ref{fig:training_curves}, we compare our approach with two baselines:

\begin{itemize} \item \textbf{VLM Preference (adapted from RLVLMF):} We follow the original prompting strategy from \cite{wang2024rl} to obtain preference labels rated as $-1$, $0$, or $1$. \item \textbf{VLM Score:} Instead of soliciting pairwise preferences, the VLM directly provides a numeric score between $0$ and $1$ for each observation. \end{itemize}

Finally, we compare our method against an \emph{oracle} that relies on ground-truth preference. VARP substantially narrows the performance gap found in standard VLM-based approaches, confirming that higher‐fidelity preference labels translate into more effective policy learning. This aligns with the earlier finding that combining sketched context with agent-aware regularization leads to better preference.

\begin{figure}
  \centering
  \begin{subfigure}{0.4\textwidth}
    \includegraphics[width=\linewidth]{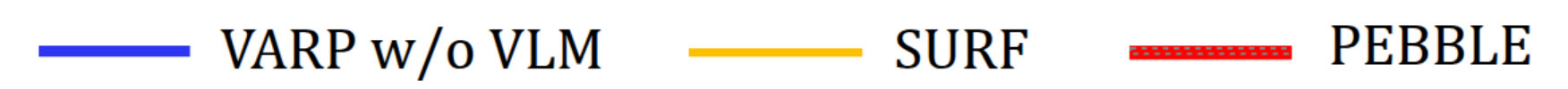}
  \end{subfigure}
  \begin{subfigure}[b]{0.48\columnwidth} % Adjust width to fit left column
    \centering
    \includegraphics[width=\linewidth]{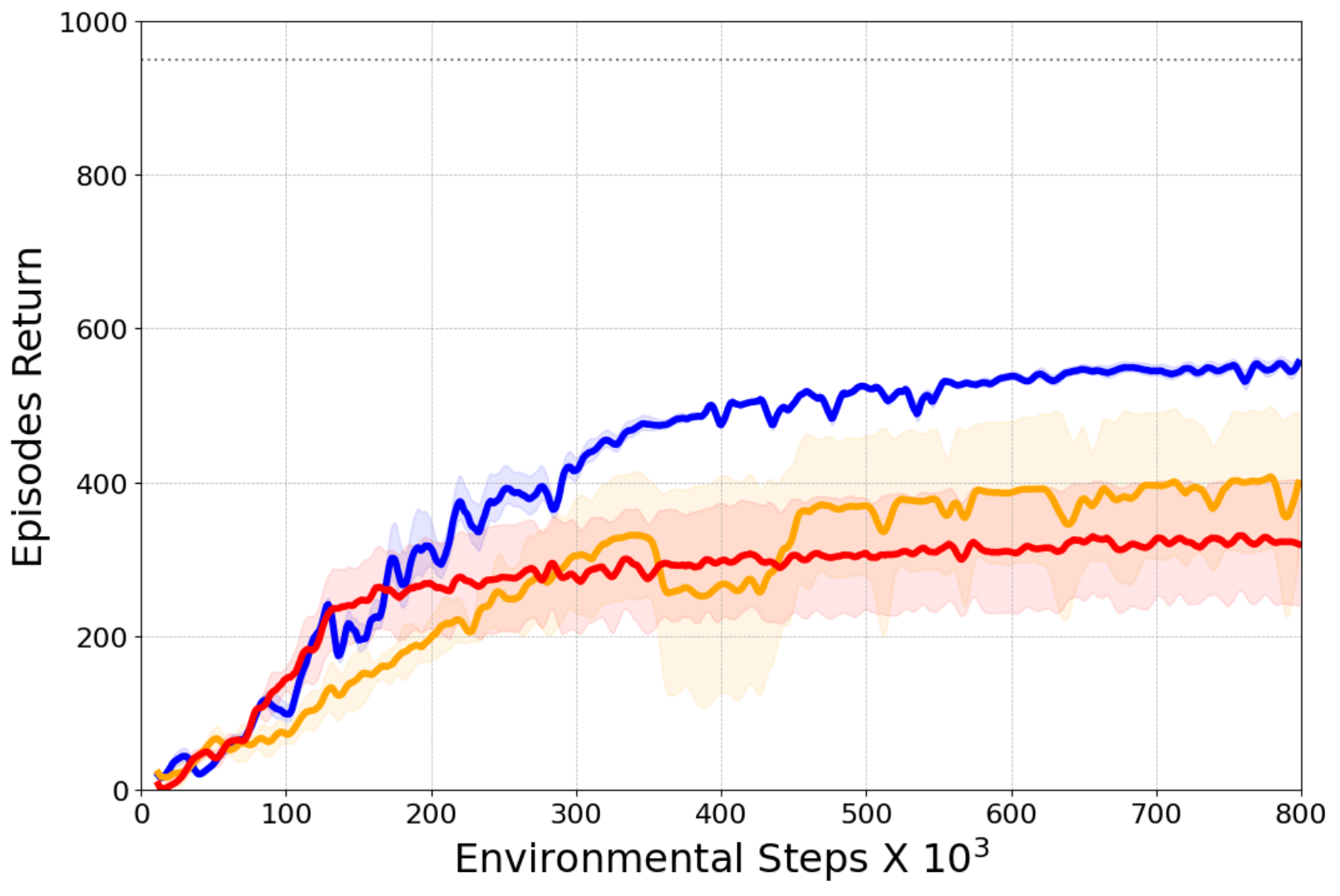}
    (a) Cheetah
    %\caption{Image 2}
  \end{subfigure}
  \hfill
  \begin{subfigure}[b]{0.48\columnwidth} % Adjust width to fit left column
    \centering
    \includegraphics[width=\linewidth]{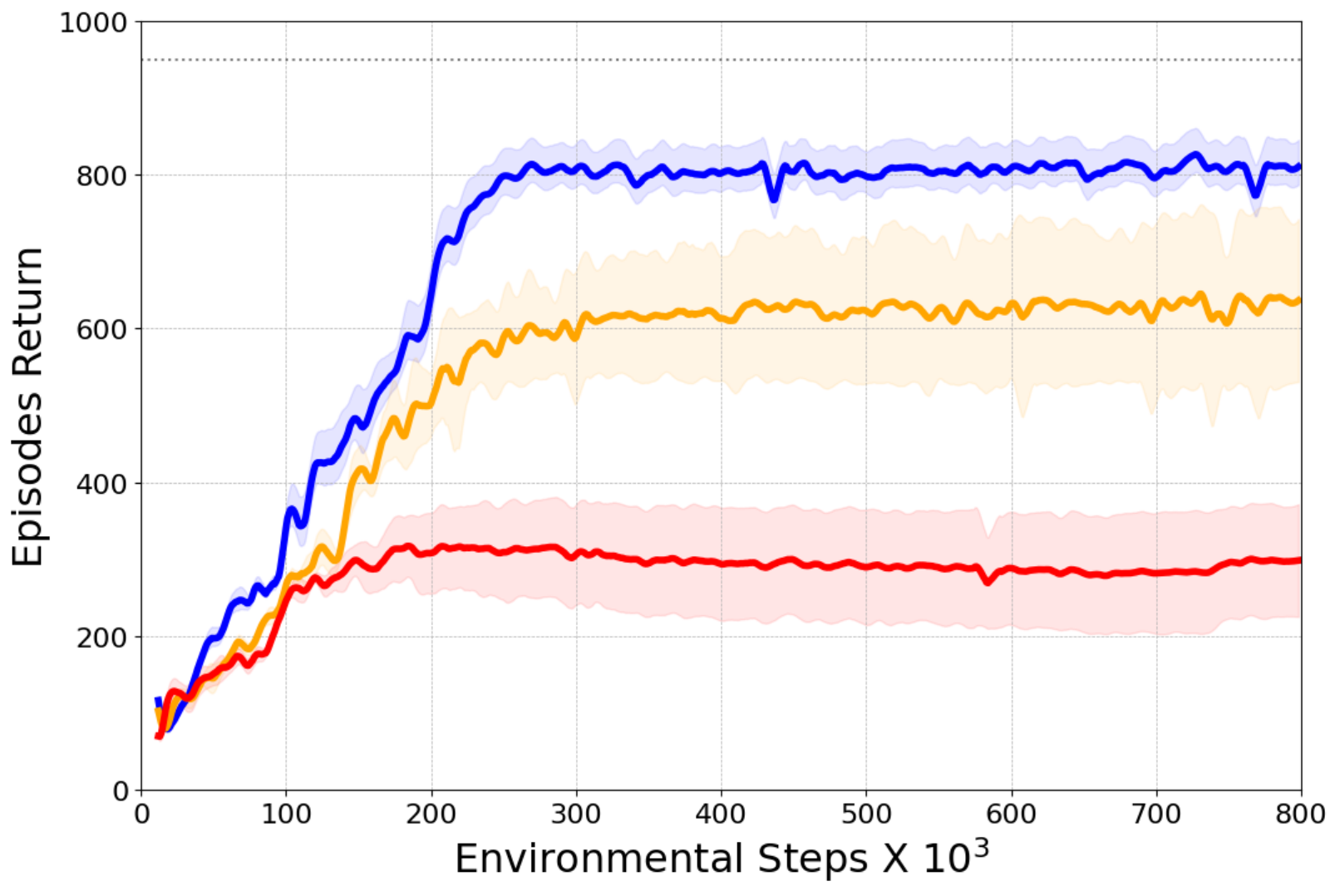}
    (a) Walker
    %\caption{Image 3}
  \end{subfigure}
  \caption{\textbf{Evaluating RLHF Accuracy with VARP in DMControl.}}
  \label{fig:rlhf_dm}
\end{figure}

\subsection{Does adding agent preference help regularize reward learning in robotic RLHF?}
Finally, we isolate the contribution of agent preference in reward learning by adding regularization. This experiment uses the scripted teacher instead of VLM that provides preferences between trajectory segments according to the true, underlying task reward as seen in the baseline methods PEBBLE~\cite{lee2021pebble} and SURF~\cite{park2022surf}. Performance is measured using episode return for DMControl tasks in Figure \ref{fig:rlhf_dm} and episode reward for Metaworld tasks in \ref{fig:rlhf_curves}. Across both these environments, we find that ignoring agent preference can lead to instability or “reward hacking,” where the policy discovers undesirable shortcuts. The results demonstrate that agent preferences provide effective regularization, leading to more stable and efficient learning compared to methods relying solely on external feedback.
For our prompting strategy, we adopt the same two-stage prompting technique as RLVLMF. 
% We include example prompts in the Appendix for reference.

\bigskip \noindent \textbf{Summary of Experimental Findings.}
Overall, our experiments indicate:
\begin{itemize} \item \textbf{Trajectory sketches} enhance the VLM’s ability to accurately compare outcomes, improving preference accuracy from approximately 68\% to 84\%. 
\item Combining \textbf{sketches and agent preference} yields significant improvements in policy performance, with episode rewards increasing by about 27\% and success rates rising from below 50\% to around 80\%. 
\item \textbf{Agent preference regularization} effectively curbs reward misalignment, reducing the misalignment metric from 0.35 to 0.15, and thus prevents reward hacking while ensuring stable convergence. 
\end{itemize}

%% file: sections/7.conclusion.tex
\section{Discussion and Conclusion}
\label{sec:discussion}

While VARP boosts preference-based reinforcement learning by integrating trajectory sketches and agent-aware regularization, a few limitations remain. First, its effectiveness is bounded by the underlying VLM, which may still mislabel complex task scenes. Generating 2D trajectory sketches adds modest overhead, particularly under occlusions or unusual camera viewpoints. Although agent preference reduces reward misalignment, tasks with extensive sub-goals or very long horizons could benefit from more granular feedback. 

In principle, one could address these issues by supplying additional intermediate images or subgoal annotations to the VLM. However, such methods would require significantly more queries and also computationally expensive— dure to larger replay buffer, potentially negating the gains from using a single final image plus a trajectory sketch. Multi-view strategies or active preference querying might mitigate these costs, but optimizing that trade-off remains an open question.

Despite these challenges, our experiments on continuous-control tasks confirm that VARP achieves more accurate preference labels and stronger policy performance compared to both VLM-only and numeric-scoring baselines. By visually portraying entire trajectories and aligning the reward with policy feasibility, VARP reduces reliance on continuous human annotation while maintaining robust, goal-aligned policy learning. We see this as a step toward scalable and interpretable preference-based RL for a wide range of robotic tasks.